\documentclass[conference]{IEEEtran}
\usepackage{times}
\usepackage[T1]{fontenc}
\usepackage[utf8]{inputenc}
\usepackage{epsfig}
\usepackage{graphicx}
\usepackage{amsmath,amsfonts,amssymb,amsthm}
\usepackage{subcaption}
\usepackage{xspace}
\usepackage{array}
\usepackage{cite}
\usepackage{booktabs}
\usepackage{multirow,multicol,blindtext}
\usepackage{microtype}
\let\labelindent\relax
\usepackage{enumitem}
\usepackage[table]{xcolor}
\usepackage[labelfont={bf}]{caption}
\usepackage{todonotes}
\usepackage{adjustbox} 
\usepackage{calc} 
\usepackage{makecell}
\usepackage{tabularx}
\usepackage{authblk}

\usepackage[numbers]{natbib}
\usepackage{hyperref}
\hypersetup{pagebackref=true, breaklinks=true, colorlinks=true, bookmarks=true}
\hypersetup{
    urlcolor   = blue,       
    linkcolor  = blue, 
    citecolor  = blue,      
    pdfborder  = {0 0 0},
}
\usepackage[capitalize]{cleveref}

\newcommand{\benchmarkname}{\texttt{navdream}\xspace}

\newcommand{\lightgray}[1]{\color{lightgray}{#1}}

\begin{document}

\title{The Constant Eye: Benchmarking and Bridging Appearance Robustness in Autonomous Driving}

\author{Jiabao Wang}
\author{Hongyu Zhou}
\author{Yuanbo Yang}
\author{Jiahao Shao}
\author{Yiyi Liao$^*$} 

\affil{Zhejiang University}

\date{}

\twocolumn[{%
    \renewcommand\twocolumn[1][]{#1}%
    \maketitle 
    
    \begin{center}
        \centering
        \includegraphics[width=\textwidth]{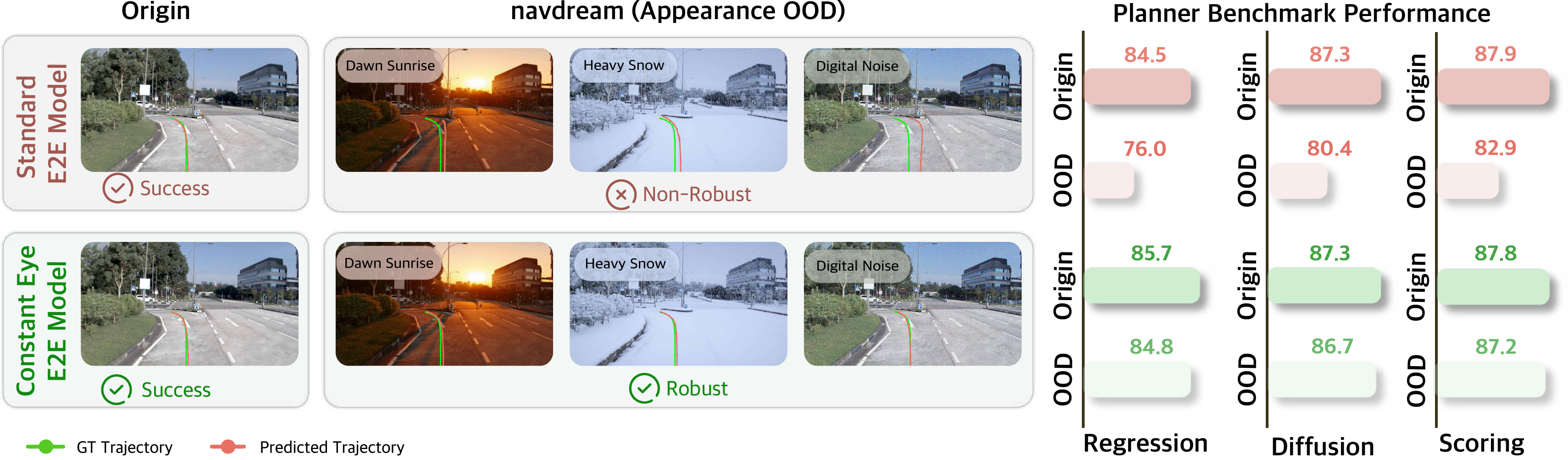}
        \captionof{figure}{\textbf{Overview}. We present \benchmarkname, a benchmark featuring out-of-distribution appearances that cause significant performance degradation across regression, diffusion, and scoring-based planning algorithms. Our method addresses this by viewing images through a ``constant eye'', leveraging a frozen visual foundation model to bridge the gap in appearance robustness.}
        \label{fig:teaser}
    \end{center}%
}]

\IEEEpeerreviewmaketitle

\begin{abstract}

Despite rapid progress, autonomous driving algorithms remain notoriously fragile under Out-of-Distribution (OOD) conditions. We identify a critical decoupling failure in current research: the lack of distinction between appearance-based shifts, such as weather and lighting, and structural scene changes. This leaves a fundamental question unanswered: Is the planner failing because of complex road geometry, or simply because it is raining? To resolve this, we establish \texttt{navdream}, a high-fidelity robustness benchmark leveraging generative pixel-aligned style transfer. By creating a visual stress test with negligible geometric deviation, we isolate the impact of appearance on driving performance. Our evaluation reveals that existing planning algorithms often show significant degradation under OOD appearance conditions, even when the underlying scene structure remains consistent. To bridge this gap, we propose a universal perception interface leveraging a frozen visual foundation model (DINOv3). By extracting appearance-invariant features as a stable interface for the planner, we achieve exceptional zero-shot generalization across diverse planning paradigms, including regression-based, diffusion-based, and scoring-based models. Our plug-and-play solution maintains consistent performance across extreme appearance shifts without requiring further fine-tuning. The benchmark and code will be made available.
\end{abstract}

\section{Introduction} 
\label{sec:introduction}

Autonomous driving (AD) holds the potential to transform transportation safety and efficiency, with End-to-End (E2E) paradigms recently emerging as a frontrunner in the field \cite{hu2023planning,jiang2023vad,chitta2022transfuser,li2024hydra}. By directly mapping sensor inputs to planning waypoints, these data-driven models bypass the complexities of traditional modular stacks \cite{7339478,9560867,DBLP:journals/corr/abs-1807-00412,DBLP:journals/corr/abs-2105-12337}. Vision-based E2E methods \cite{hu2023planning,weng2024drive,wozniak2025prixlearningplanraw,li2025generalized}, in particular, have gained immense traction due to the scalability and semantic richness of camera sensors. However, despite their impressive results, these models remain notoriously fragile. In out-of-distribution (OOD) or long-tailed scenarios, even state-of-the-art systems often suffer from severe performance degradation, posing a significant barrier to reliable real-world deployment \cite{chen2024end}.

While OOD challenges are multi-faceted, appearance shifts are a critical factor for vision-based models, as changes in lighting or weather can drastically alter input pixels even without changing the physical road layout. From a probabilistic perspective, a driving scenario is a joint distribution of appearance and scene geometry. Yet, current research \cite{wang2024drive,pan2024vlp,li2024llada,caesar2020nuscenes} typically entangles these dimensions. For example, evaluating on a rainy dataset often involves a simultaneous shift in geography, making it impossible to determine whether a planning failure stems from a complex new topology or a breakdown in visual robustness. Is the planner failing because the road geometry is complex, or simply because it is raining? Without a controlled way to factorize these variables, this fundamental question remains unanswered.

To this end, we first establish \benchmarkname, a high-fidelity robustness evaluation benchmark leveraging a state-of-the-art generative model, Flux \cite{labs2025flux1kontextflowmatching}. By performing pixel-aligned style transfer on real-world sequences from the NAVSIM \cite{Dauner2024NEURIPS} dataset, we synthesize a wide spectrum of OOD environments with negligible deviation in semantic and geometric information. This methodology provides a first-of-its-kind ``visual stress test'', isolating appearance shifts from structural factors to precisely quantify whether a driving system can maintain planning stability when only the visual domain undergoes substantial changes.
 
Evaluating representative E2E driving models \cite{chitta2022transfuser,liao2025diffusiondrive,li2025generalized} on \benchmarkname reveals their fragility. 
While some robustness-oriented methods \cite{tobin2017domain,wang2021versatile,osinski2020simulation} train on additional domains and achieve improved performance, they still struggle to generalize to unseen domains.
This suggests that existing methods over-rely on appearance cues instead of anchoring decisions in stable scene semantics and geometry.

To address this, we explore a simple yet effective paradigm that anchors end-to-end planning on a frozen foundation model, effectively viewing the world through a ``constant eye'' to ensure appearance robustness. 
By integrating a pre-trained DINOv3 backbone, we leverage self-supervised features that possess inherent appearance robustness, capturing consistent semantic priors that remain invariant to appearance changes.
This approach provides a plug-and-play solution compatible with multiple E2E architectures, including regression \cite{chitta2022transfuser}, diffusion \cite{liao2025diffusiondrive}, and scoring-based \cite{li2025generalized} models. Experimental results demonstrate that this paradigm achieves exceptional zero-shot generalization across diverse visual appearances without requiring any target-domain training, while preserving competitive performance on the original NAVSIM benchmarks~\cite{Dauner2024NEURIPS,Cao2025CORL}. Our contributions are as follows:
\begin{itemize}[leftmargin=*,topsep=2pt,itemsep=1pt]

\item We examine the independent impact of purely visual domain shifts on E2E planning stability via a novel benchmark, \benchmarkname, leveraging an advanced generative model for pixel-aligned style transfer. \benchmarkname provides a visual stress test by synthesizing diverse appearance variations with negligible geometric deviation, enabling robustness evaluation under severe appearance shifts.

\item We propose a simple yet effective solution to enhance the appearance robustness via a frozen vision foundation model, DINOv3, as a standardized perception interface, serving as a plug-and-play module across diverse planning paradigms. 

\item Experimental results demonstrate that our proposed solution achieves exceptional zero-shot generalization on \benchmarkname across regression-based, diffusion-based, and scoring-based planners without any target-domain training, while retaining good performance on the original NAVSIM benchmarks.

\end{itemize}

\section{Related Work}  
\label{sec:rel_work}

\subsection{End-to-End Autonomous Driving}
\label{sec:related_work_e2e}
E2E-AD aims to directly map raw sensory inputs to control commands or trajectory waypoints. Early research primarily focused on \textit{regression-based planning}~\cite{chitta2022transfuser,hu2023planning,jiang2023vad,weng2024drive}. For instance, UniAD \cite{hu2023planning} stands as a pioneering work that integrates multiple perception tasks into a unified and differentiable framework. VAD \cite{jiang2023vad} leverages vectorized scene representation for planning, while PARA-Drive \cite{weng2024drive} achieves high efficiency by performing mapping, planning, motion prediction, and occupancy prediction tasks in parallel. %
However, since real-world driving scenarios are inherently multi-modal \cite{chen2024vadv2}, deterministic regression models often struggle with ambiguous labels.
\textit{Diffusion-based} methods tackle this issue by modeling complex trajectory distributions \cite{liao2025diffusiondrive,zou2025diffusiondrivev2,jiang2025transdiffuser,xing2025goalflow} inspired by Diffusion Policy \cite{chi2025diffusion}. For example, DiffusionDrive \cite{liao2025diffusiondrive} proposes a truncated diffusion policy that begins the denoising process from an anchored Gaussian distribution. 
Another direction involves \textit{scoring-based methods} \cite{li2025hydra,li2025generalized,yao2025drivesuprim,li2025end}, which evaluate a trajectory vocabulary and select the optimal one. By shifting from local regression to global selection, Scoring-based methods can better handle diverse driving maneuvers and even bypass the limitations of suboptimal expert behaviors \cite{li2025ztrs}.

Despite these advances in planning paradigms, the robustness of the underlying visual backbones remains a critical bottleneck. Most existing E2E models rely on backbones that are highly susceptible to visual domain shifts. In this work, we provide a simple yet effective solution to enhance appearance robustness via a frozen visual foundation model, DINOv3, demonstrating its superior robustness across diverse planning paradigms.

\subsection{Domain Adaptation in Autonomous Driving}
\label{sec:related_work_domain_adaptation}
Domain adaptation (DA) is essential to bridge the gap between training environments and diverse real-world scenarios in autonomous driving. Domain randomization \cite{tobin2017domain} is a simple and effective technique by synthesizing training samples to cover a broader distribution to improve robustness \cite{wang2021versatile,osinski2020simulation}. More sophisticated methods \cite{xing2021domain,pan2024vlp,yasarla2025roca} employ feature-level alignment to extract domain-invariant representations. However, these reactive adaptation strategies inherently require prior exposure to target-domain data and often necessitate additional retraining or fine-tuning stages, which limits their scalability in handling the infinite visual permutations of the open world. This challenge has prompted a shift toward leveraging Vision Foundation Models (VFMs) as a more scalable alternative that provides inherently robust representations learned from internet-scale data without requiring domain-specific adaptation.

\begin{figure*}[t]
    \centering
    \begin{tabular}{c | l}
        \begin{minipage}[c]{0.16\textwidth}
            \centering
            \begin{subfigure}[b]{\textwidth}
                \centering
                \includegraphics[width=\textwidth]{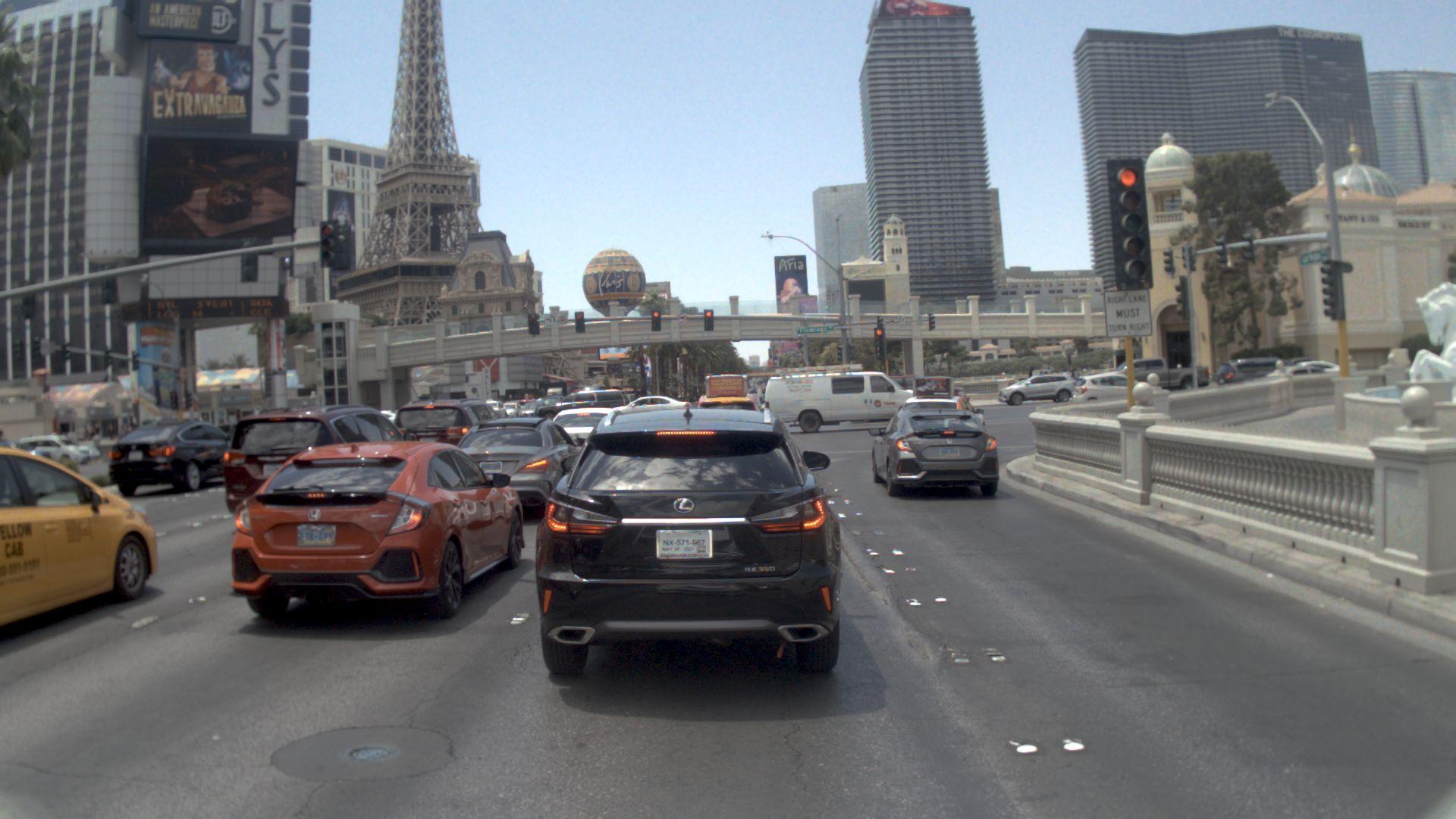}
                \caption*{Origin}
            \end{subfigure}
        \end{minipage}
        &
        \begin{minipage}[c]{0.80\textwidth}
            \centering
            \begin{subfigure}[b]{0.19\linewidth}
                \centering
                \includegraphics[width=\textwidth]{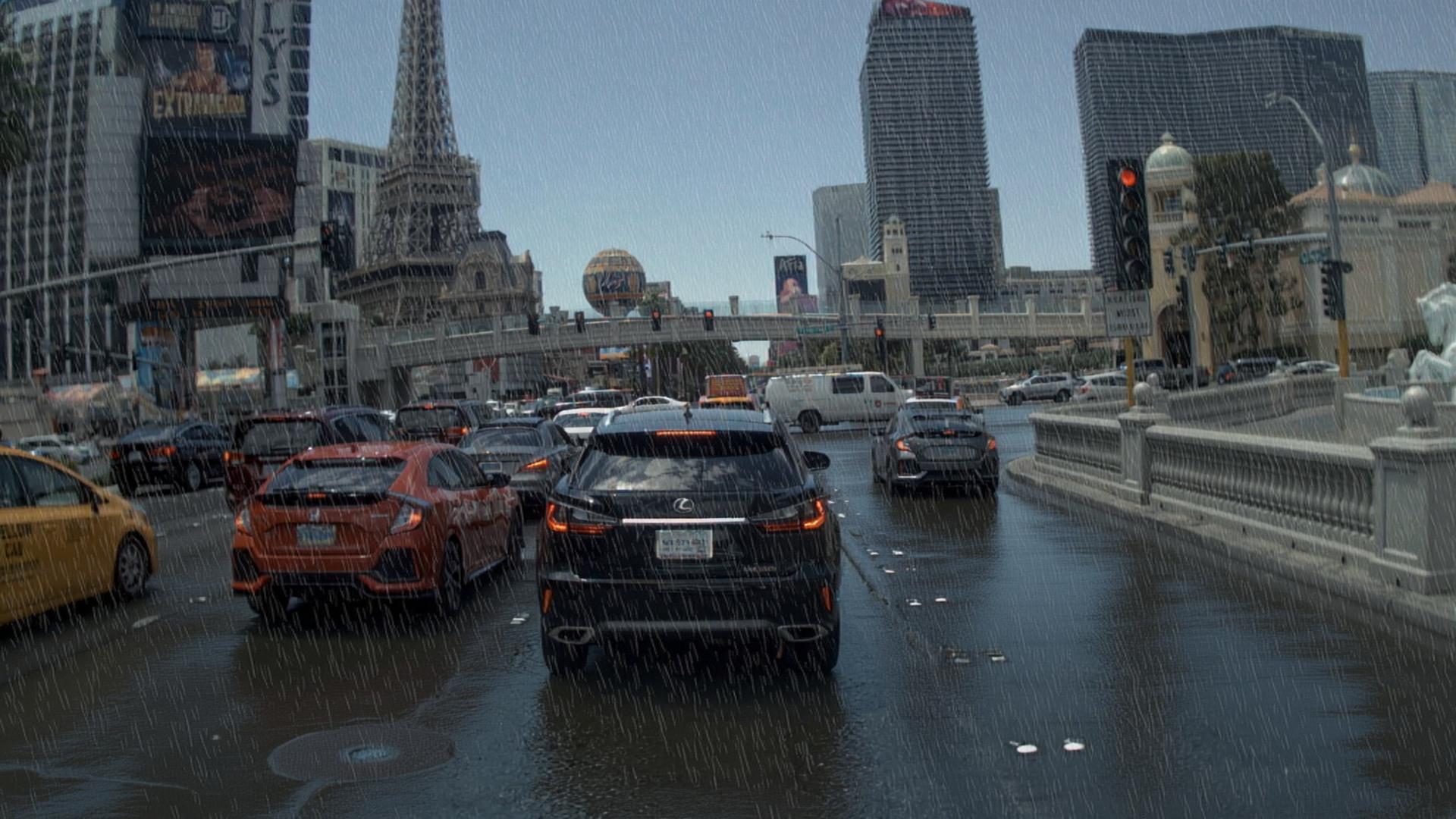}
                \caption*{Heavy Rain}
            \end{subfigure}
            \hfill
            \begin{subfigure}[b]{0.19\linewidth}
                \centering
                \includegraphics[width=\textwidth]{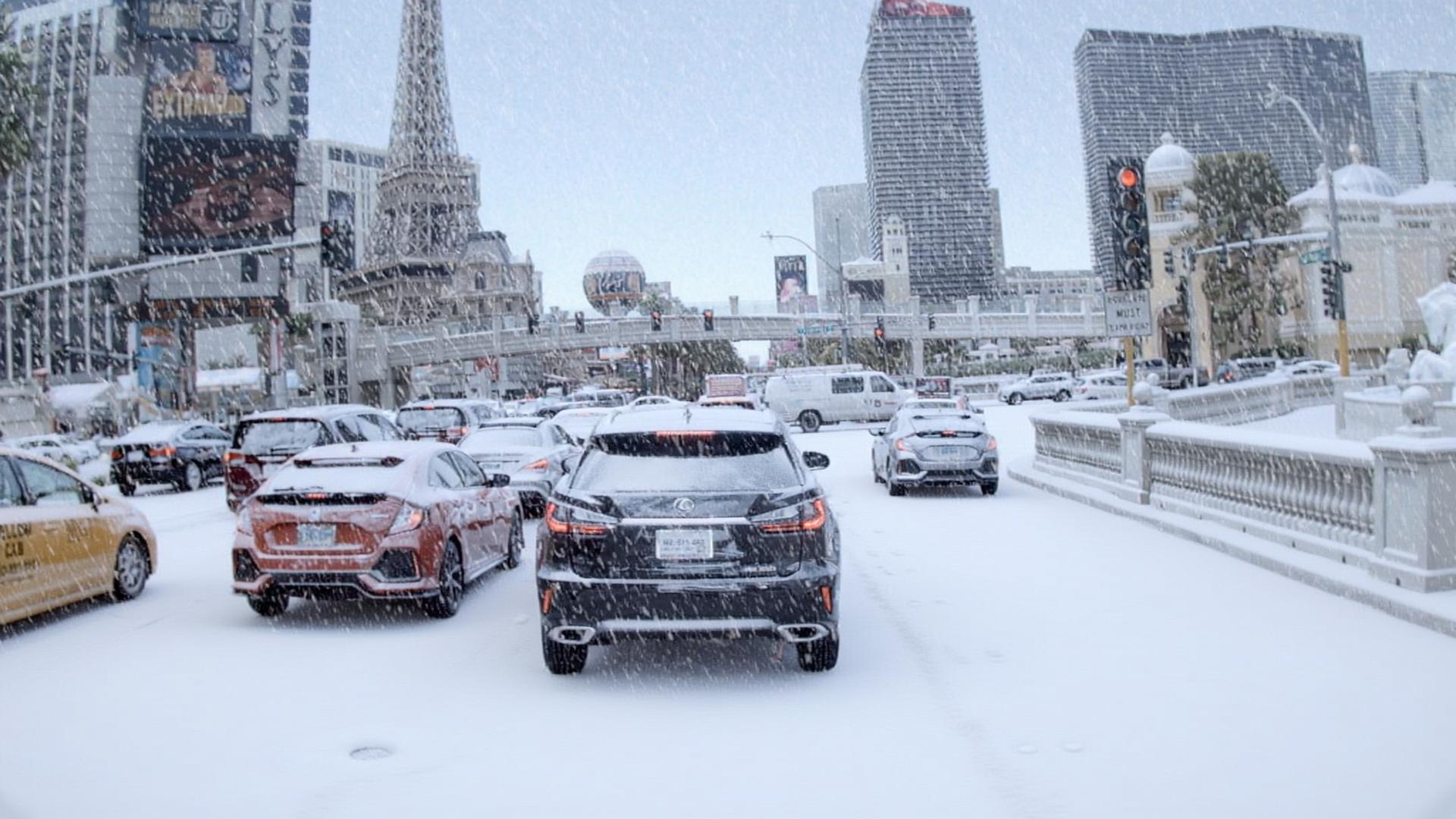}
                \caption*{Heavy Snow}
            \end{subfigure}
            \hfill
            \begin{subfigure}[b]{0.19\linewidth}
                \centering
                \includegraphics[width=\textwidth]{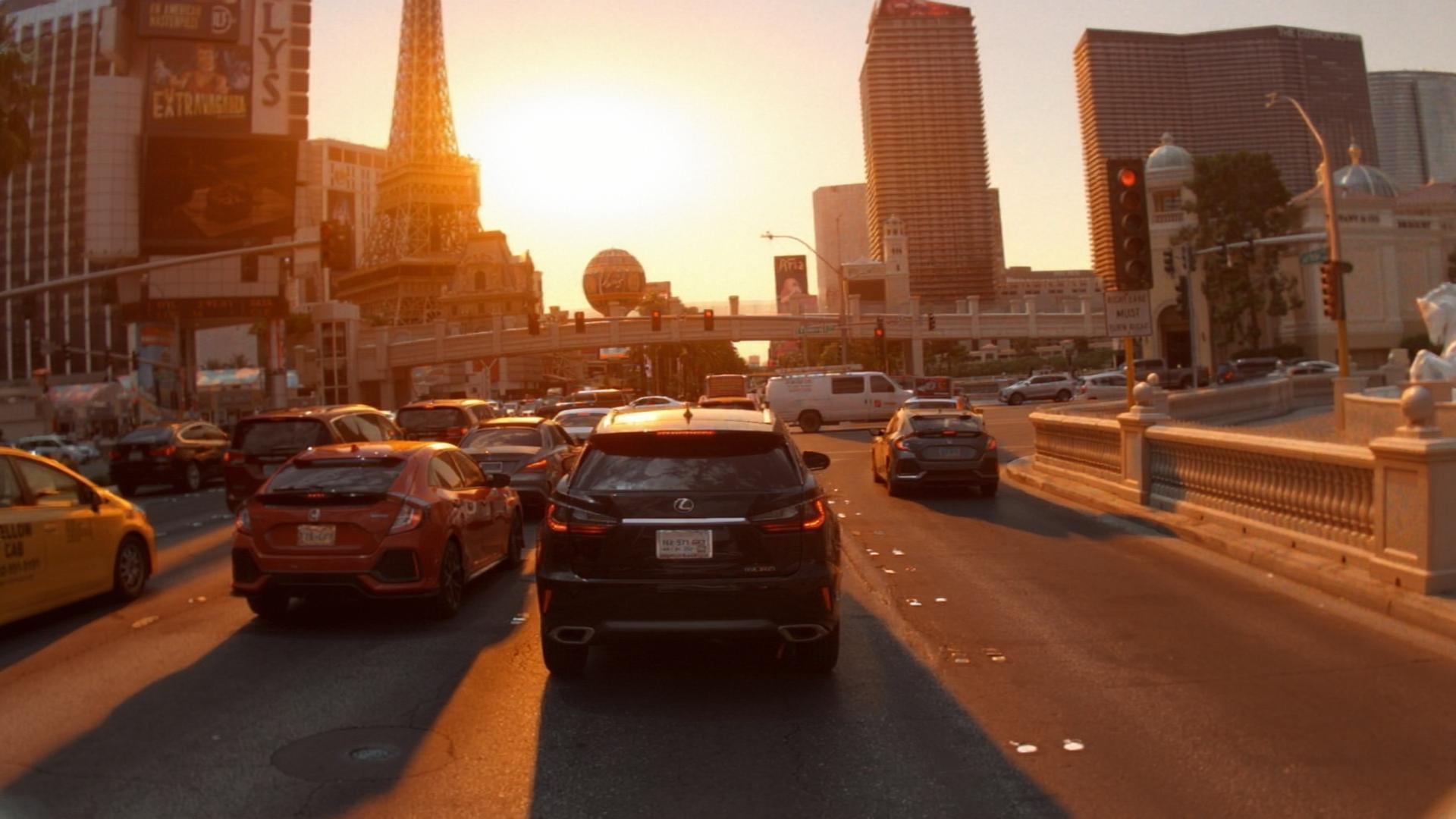}
                \caption*{Dawn Sunrise}
            \end{subfigure}
            \hfill
            \begin{subfigure}[b]{0.19\linewidth}
                \centering
                \includegraphics[width=\textwidth]{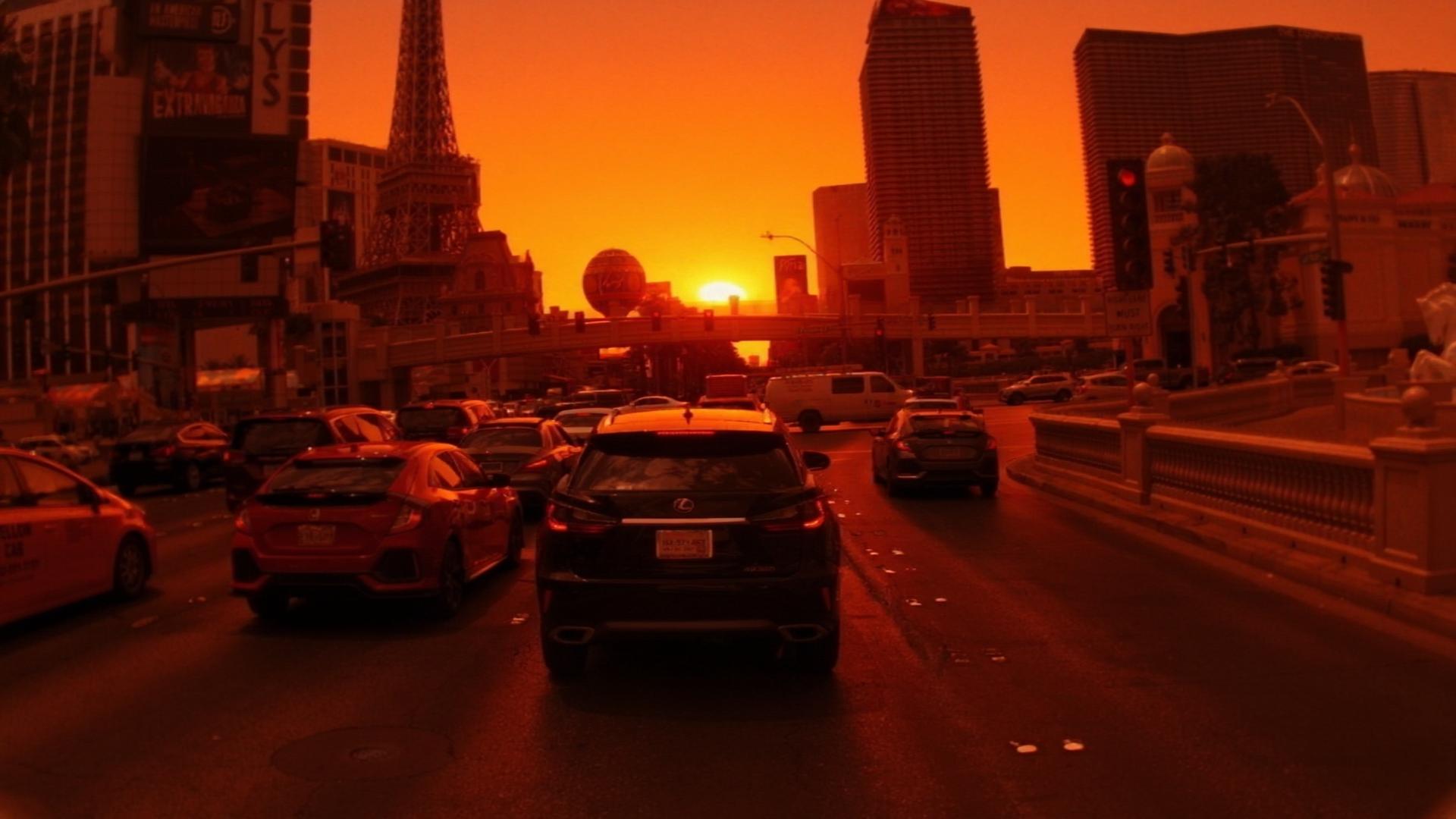}
                \caption*{Dusk Sunset}
            \end{subfigure}
            \hfill
            \begin{subfigure}[b]{0.19\linewidth}
                \centering
                \includegraphics[width=\textwidth]{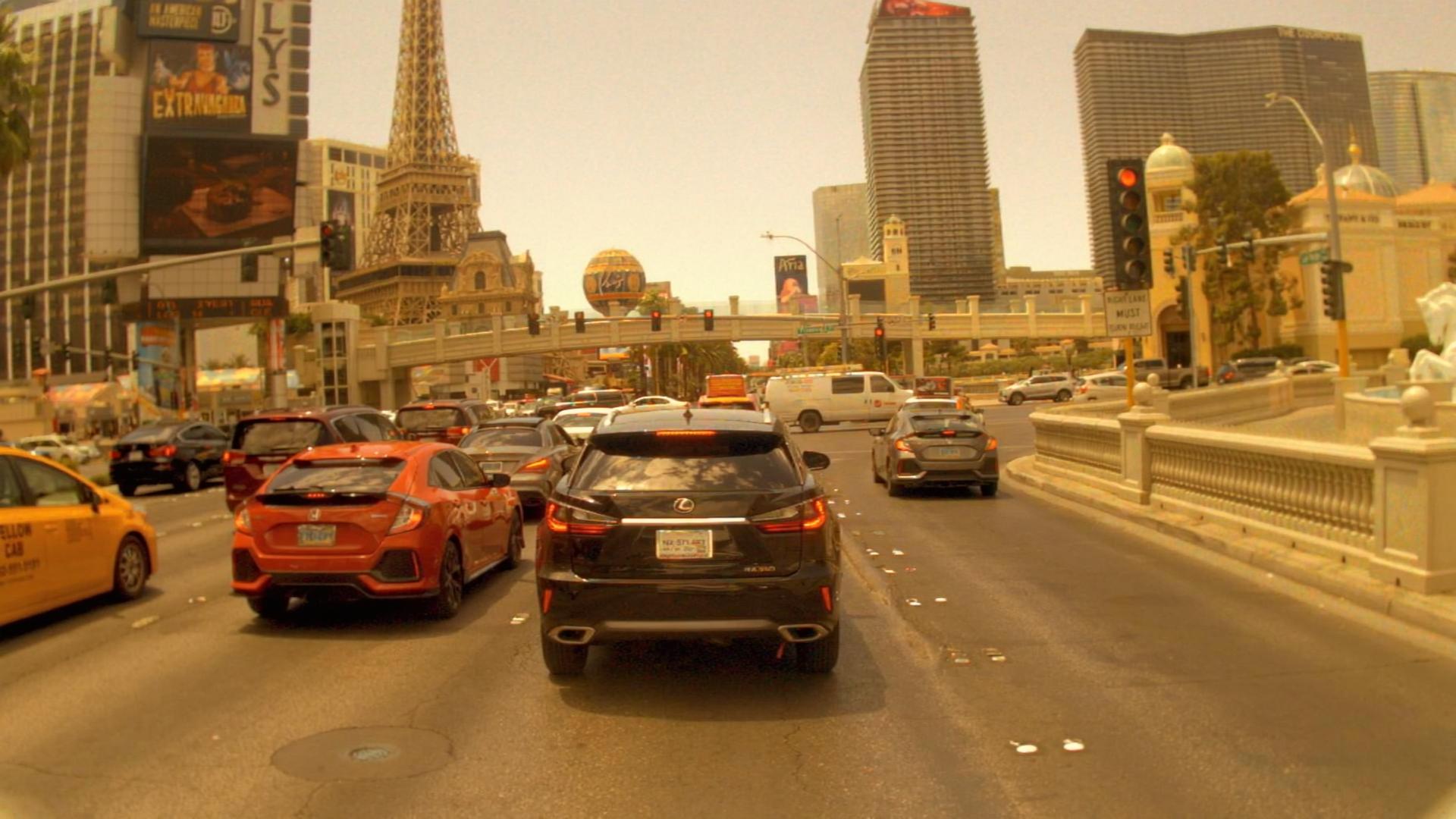}
                \caption*{Light Dust}
            \end{subfigure}

            \vspace{2mm}

            \begin{subfigure}[b]{0.19\linewidth}
                \centering
                \includegraphics[width=\textwidth]{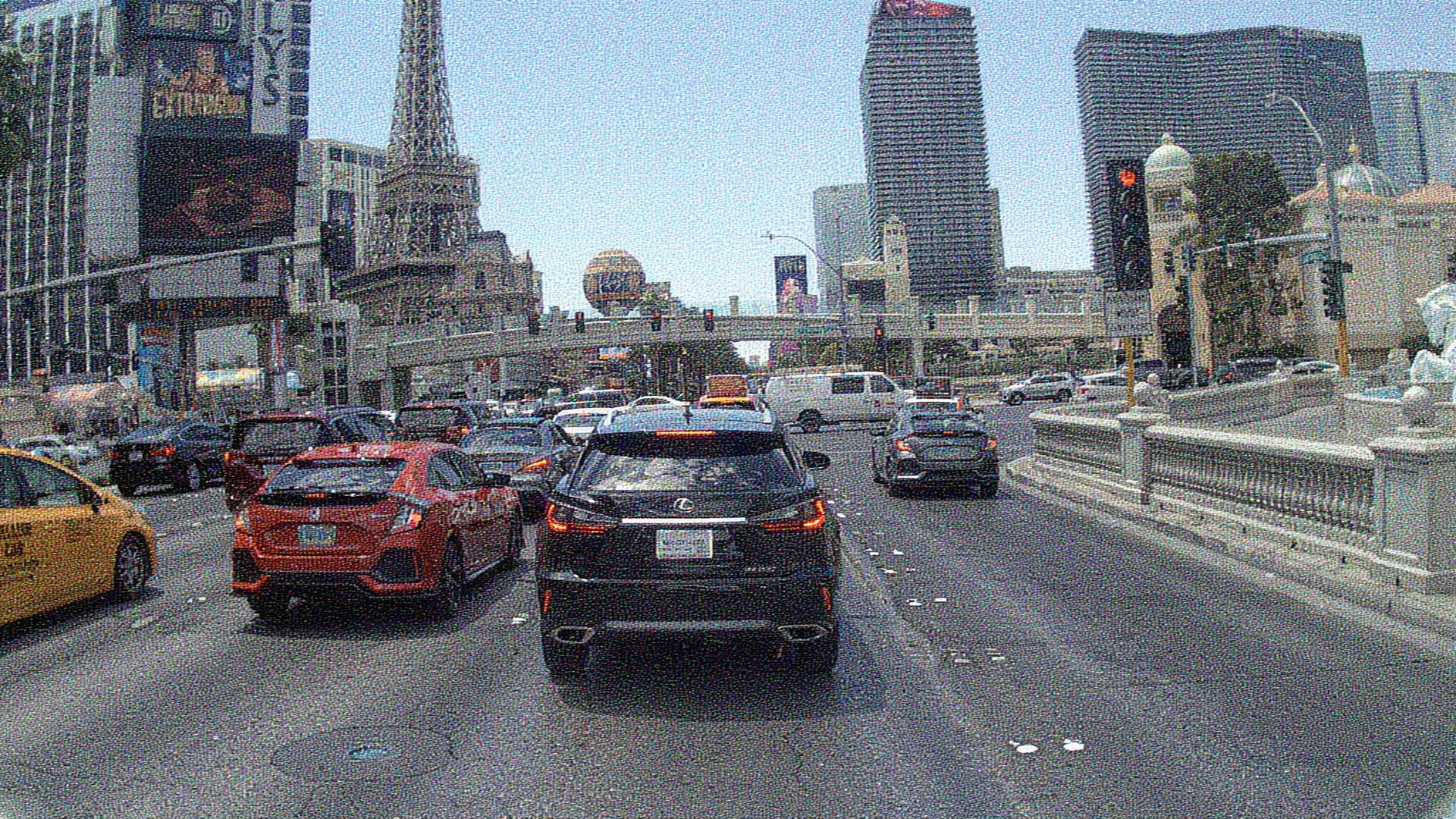}
                \caption*{Digital Noise}
            \end{subfigure}
            \hfill
            \begin{subfigure}[b]{0.19\linewidth}
                \centering
                \includegraphics[width=\textwidth]{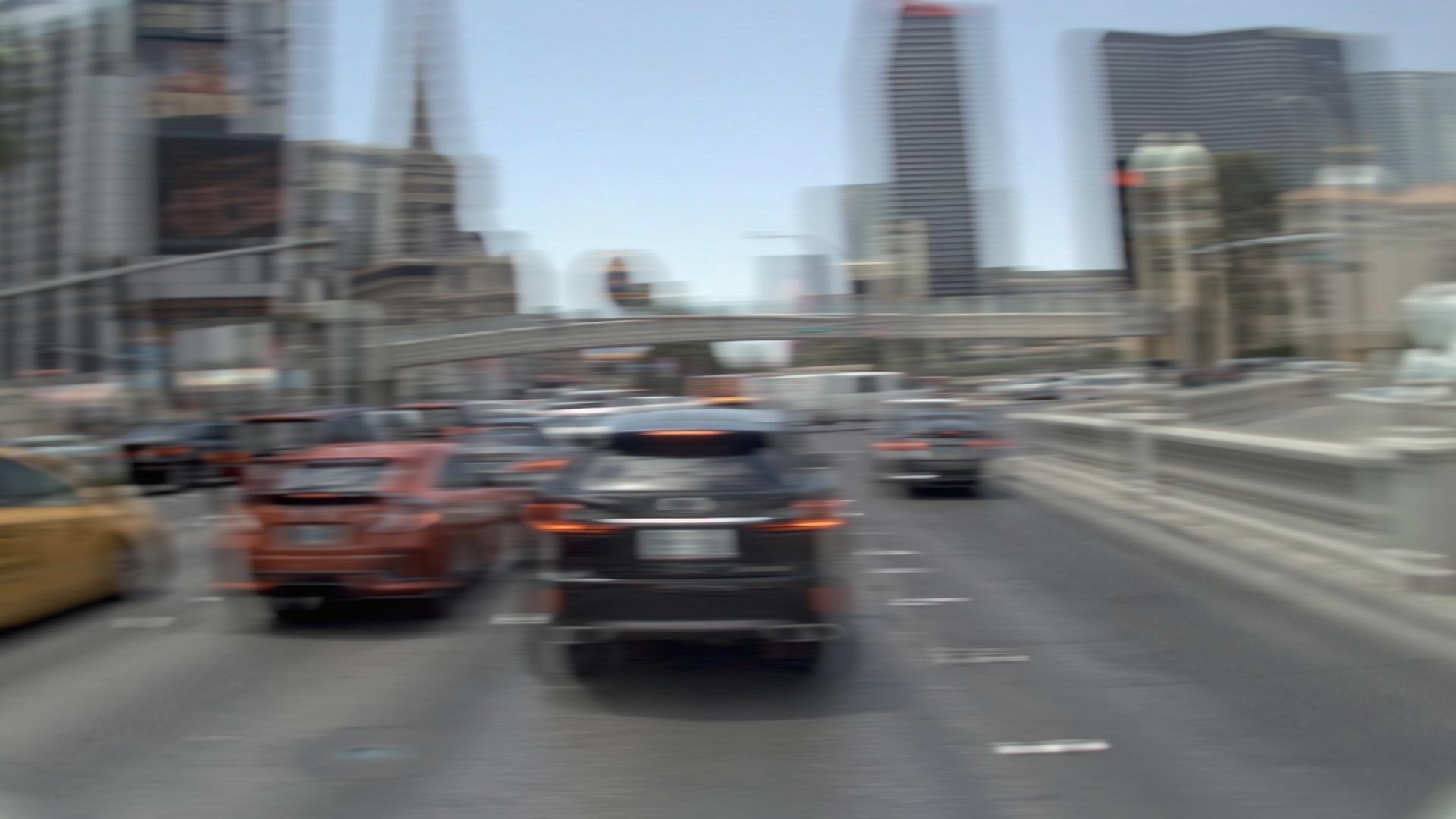}
                \caption*{Motion Blur}
            \end{subfigure}
            \hfill
            \begin{subfigure}[b]{0.19\linewidth}
                \centering
                \includegraphics[width=\textwidth]{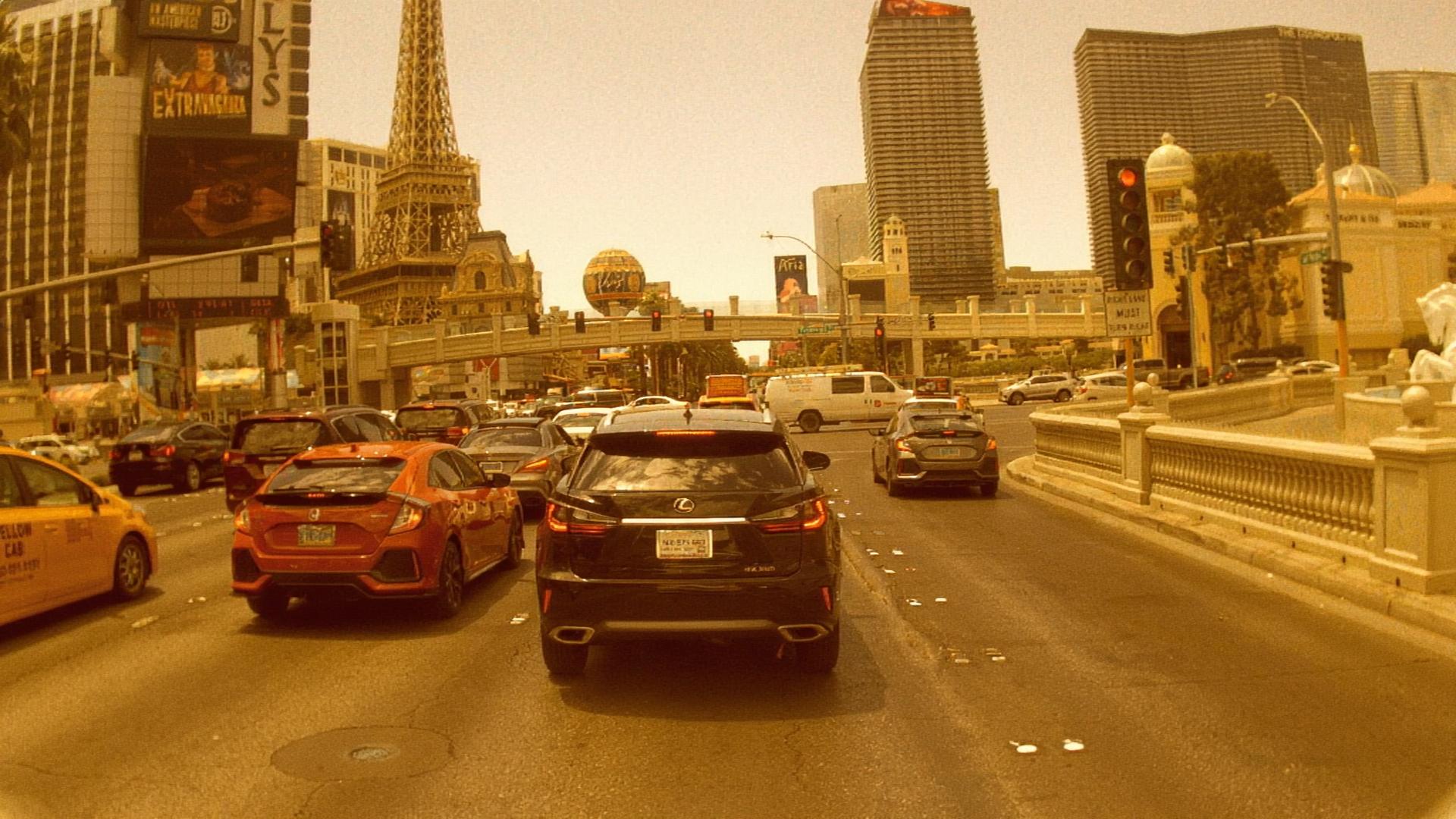}
                \caption*{Vintage Photo}
            \end{subfigure}
            \hfill
            \begin{subfigure}[b]{0.19\linewidth}
                \centering
                \includegraphics[width=\textwidth]{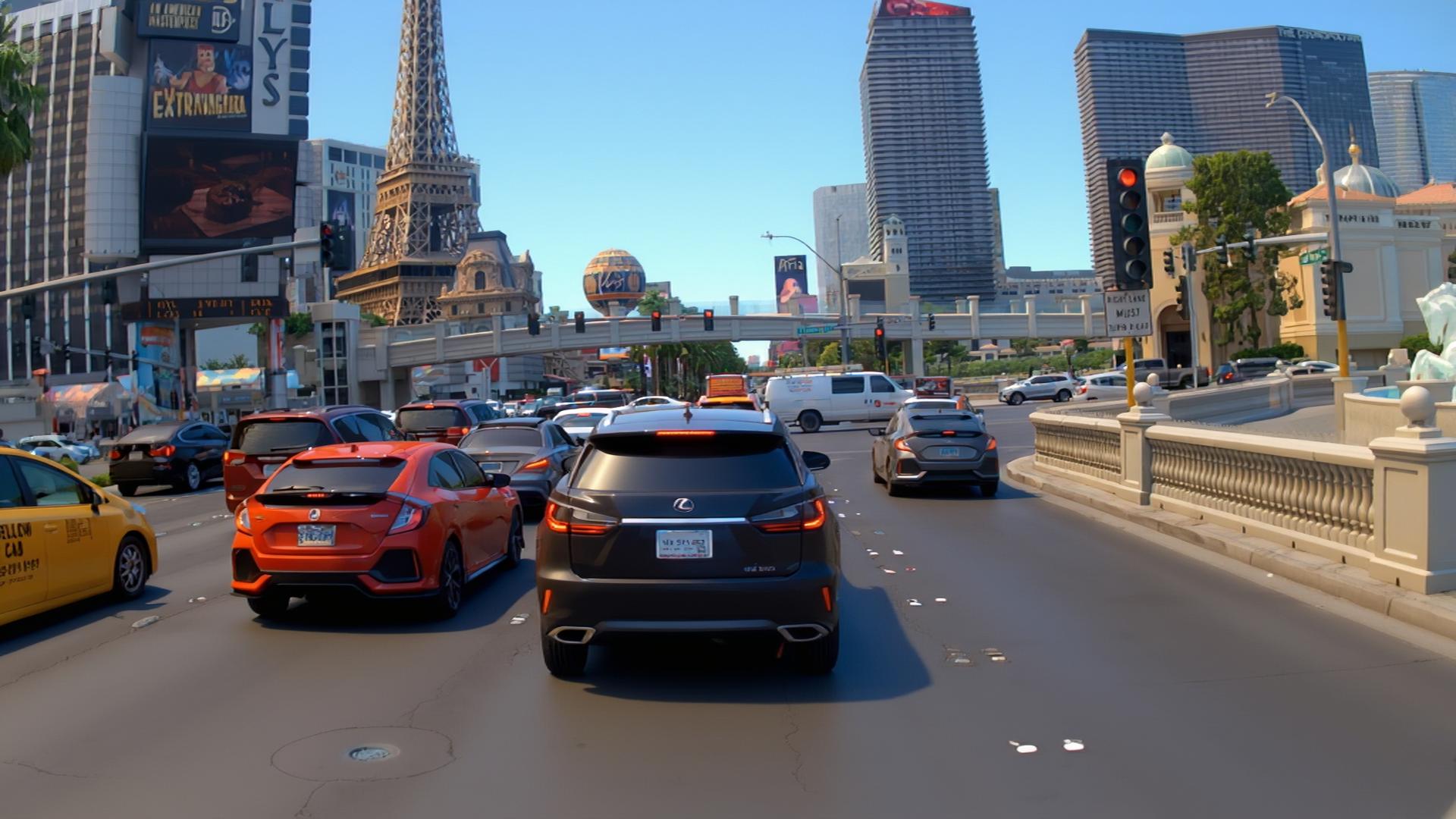}
                \caption*{CARLA Toy}
            \end{subfigure}
            \hfill
            \begin{subfigure}[b]{0.19\linewidth}
                \centering
                \includegraphics[width=\textwidth]{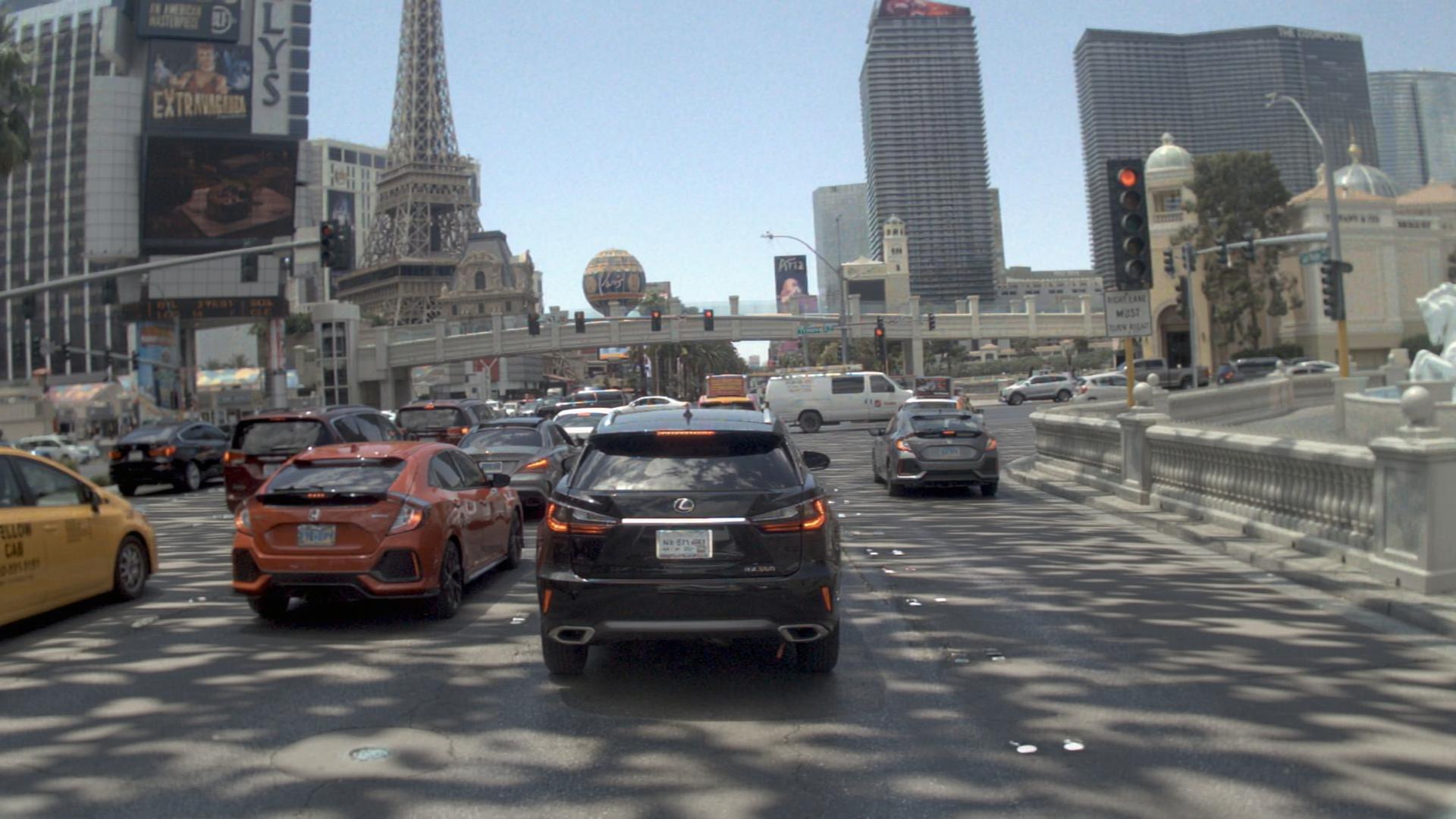}
                \caption*{Dappled Light}
            \end{subfigure}
        \end{minipage}
    \end{tabular}
    \caption{\textbf{Visual taxonomy of the appearance-based OOD shifts in \benchmarkname.} We illustrate the original frame alongside 10 synthesized stylistic variations generated by the Flux model. All transformations preserve the underlying 3D geometry and semantic structures while shifting the visual domain to various OOD conditions.}
    \label{fig:benchmark_taxonomy}
\end{figure*}

\subsection{Vision Foundation Model in Autonomous Driving}
\label{sec:related_work_vfm}
Vision Foundation Models (VFMs) including DINO~\cite{simeoni2025dinov3}, SigLIP~\cite{zhai2023sigmoidlosslanguageimage}, and SAM~\cite{carion2025sam} have matured significantly and are now extensively utilized across a wide range of perception tasks in autonomous driving \cite{liu2024grounding,yang2024depth,feng2025vipocc}, demonstrating remarkable universality and robustness. In the context of the planning task, FROST-Drive~\cite{dong2026frost} has explored the integration of frozen vision encoders from large-scale VLMs to leverage their rich world knowledge for more robust trajectory generation. Similarly, some works \cite{wang2024drive,mallak2026lessdrivebettergeneralizable} have explored utilizing multi-modal vision encoders within the planning pipeline to strengthen resilience against OOD scenarios. However, these existing approaches are typically confined to a specific architecture or a single planning paradigm. 
In contrast, our work places a greater emphasis on the systematic analysis of appearance-based robustness, positioning the frozen DINOv3 backbone as a plug-and-play, universal component that serves as a standardized perception interface for diverse E2E-AD frameworks.

\section{navdream Benchmark}
\label{sec:Benchmark}

\subsection{Decoupling Appearance and Geometry}
\label{sec:decoupling_appearance_geometry}

The primary objective of our benchmark, referred to as \benchmarkname, is to isolate the impact of visual appearance from structural scene factors. To provide a formal basis for this decoupling, we model a driving scenario $S$ as a joint distribution of its visual appearance $A$ and underlying geometry $G$. Following the chain rule of probability, this distribution can be factorized as: 
\begin{equation}
    P(S) = P(A, G) = P(A|G)P(G),
\end{equation} where $G$ represents the invariant structural elements and $P(A|G)$ denotes the conditional visual representation given a fixed geometry. Unlike traditional OOD evaluations that typically involve simultaneous shifts in both $P(G)$ and $P(A|G)$, \benchmarkname adopts a rigorous variable isolation approach. Specifically, for any given scene, we fix $P(G)$ to ensure that the ground-truth expert trajectory and the structural scene factors remain identical across all visual variations. We then systematically vary $P(A|G)$ by synthesizing 10 distinct appearance styles (plus 1 original reference). This setup enables a ``pure visual attack'' on the perception layer, ensuring that any resulting planning instability is explicitly attributed to changes in appearance rather than shifts in scene semantics.

\subsection{Scenario Synthesis and Appearance Taxonomy}
\label{sec:scenario_synthesis_taxonomy}
We build \benchmarkname upon the NAVSIM \cite{Dauner2024NEURIPS} dataset. NAVSIM is a real-world planning-oriented dataset built upon OpenScene \cite{contributors2023openscene}, a compact redistribution of nuPlan \cite{karnchanachari2024towards}, which deliberately excludes trivial situations such as stationary scenes or constant speed driving, to focus on evaluating E2E-AD systems in complex urban environments. \par

In \benchmarkname, we choose the \texttt{navtest} split of NAVSIM as the source domain. To synthesize the appearance-based OOD variations, we employ the Flux \cite{labs2025flux1kontextflowmatching} generative model to perform image-to-image style transfer on the original camera frames. The generation process leverages Flux’s inherent capacity to maintain spatial consistency during style translation. By utilizing tailored text prompts, we transform the visual context with a negligible shift in semantic and geometric structures. This approach ensures that the underlying geometry $G$ is preserved, as defined in \cref{sec:decoupling_appearance_geometry}. We define a taxonomy of 10 appearance shifts, including: Heavy Rain, Heavy Snow, Dawn Sunrise, Dusk Sunset, Light Dust, Vintage Photo, Digital Noise, Motion Blur, CARLA Toy, and Dappled Light. The visual results of these transformations are illustrated in \cref{fig:benchmark_taxonomy}. In total, \benchmarkname comprises 2,304 base scenarios, each synthesized with 10 distinct visual variations, resulting in a comprehensive collection of 23,040 augmented evaluation cases. Accounting for the multi-view sensor configuration (front-left, front, and front-right camera views), the benchmark encompasses a total of 69,120 high-resolution images. The entire synthesis process incurred a computational overhead of approximately 960 GPU hours.

\begin{figure*}[t]
    \centering
    \includegraphics[width=\textwidth]{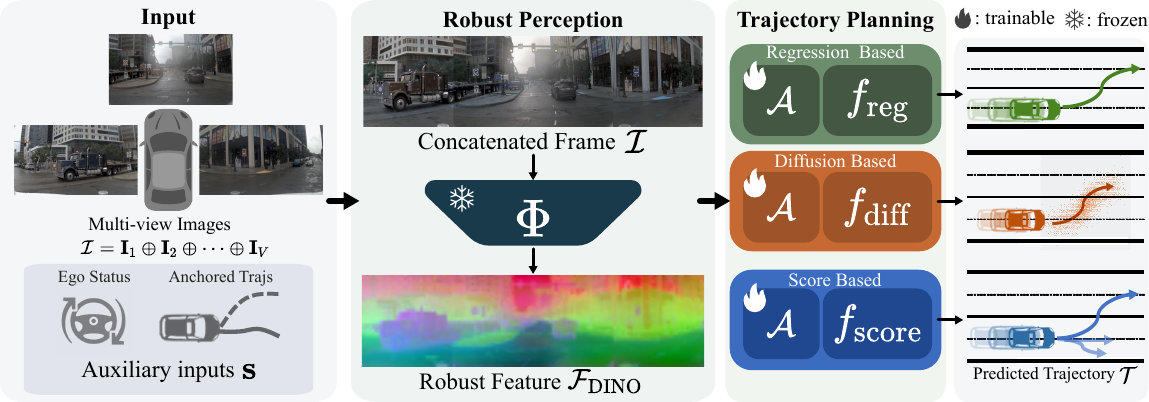}
    \caption{\textbf{Method.} We utilize a frozen DINOv3 backbone $\Phi$ to extract features from raw camera inputs and these features maintain consistent semantic information across visual domains. These structural representations are then processed by a lightweight feature adapter $\mathcal{A}$ to reduce dimensionality. This plug-and-play solution can be integrated into regression, diffusion, and scoring-based planning paradigms to ensure robust trajectory generation across varying appearance conditions.}
    \label{fig:method_pipeline}
\end{figure*}

\subsection{Evaluation Metrics}
\label{sec:evaluation_metrics}
We adopt the Extended Predictive Driver Model Score (EPDMS) \cite{Cao2025CORL} as our primary evaluation metric. EPDMS offers a comprehensive assessment with several sub-metrics:
\begin{equation}
\text{EPDMS} = \underbrace{\prod_{m \in \mathcal{M}_{\text{pen}}} S_m}_{\text{penalty terms}} \cdot \underbrace{\frac{\sum_{m \in \mathcal{M}_{\text{avg}}} w_m S_m}{\sum_{m \in \mathcal{M}_{\text{avg}}} w_m}}_{\text{weighted average terms}},
\end{equation}
where $S_m$ is the sub-metric: the penalty-term set $\mathcal{M}_{\text{pen}}$ includes No-at-fault Collisions (NC), Drivable Area Compliance (DAC), Driving Direction Compliance (DDC), and Traffic Light Compliance (TLC); the weighted-average set $\mathcal{M}_{\text{avg}}$ includes Time-to-Collision (TTC), Ego Progress (EP), Lane Keeping (LK), History Comfort (HC), and extended comfort (EC). Furthermore, a human filter is incorporated: a penalty is ignored if the same rule violation is committed by the human expert in the identical scenario. Note that we adopt the official bug-fixed version of EPDMS.

\section{Method}  
\label{sec:Method}

In this section, we investigate the integration of a frozen DINOv3 backbone \cite{simeoni2025dinov3} into End-to-End Autonomous Driving, aiming to extract semantic and structural representations that are robust to visual appearance variations. Our primary objective is to demonstrate the applicability of this consistent representation across diverse planning paradigms. The overall pipeline of our framework is illustrated in \cref{fig:method_pipeline}.

\subsection{Problem Formulation}
\label{subsec:Problem_Formulation}
The objective of an E2E-AD system is to learn a policy $\pi$ that maps raw visual observations $\mathcal{I}$ to a safe and feasible future trajectory $\mathcal{T}$. Specifically, we consider a multi-camera setup where the cropped surrounding views are horizontally concatenated as $
\mathcal{I} = \mathbf{I}_1 \oplus \mathbf{I}_2 \oplus \cdots \oplus \mathbf{I}_V$,
where $\oplus$ denotes horizontal concatenation. The planning task is formulated as a functional composition $\pi = f(\cdot, \mathbf{s}) \circ \Phi$:
\begin{equation}
\mathcal{T} = f(\Phi(\mathcal{I}), \mathbf{s}),
\end{equation}
where $\Phi$ denotes the visual backbone that extracts latent visual features and $f$ represents the planning head. Here, $f$ generates the future trajectory $\mathcal{T}$ by integrating visual features with auxiliary inputs $\mathbf{s}$ (e.g., ego-vehicle status or pre-defined anchors). The resulting trajectory is represented as a sequence of discrete waypoints $\mathcal{T} = \{ \mathbf{w}_t \}_{t=1}^T$ in the ego-coordinate system.

\subsection{Perception with Frozen DINOv3}
\label{subsec:perception_frozen_dinov3}
Rather than employing the task-specific supervised backbones typical of end-to-end architectures, we utilize a frozen, self-supervised DINOv3 \cite{simeoni2025dinov3} as the core visual encoder $\Phi$. The visual input $\mathcal{I}$ is transformed into a feature grid:
\begin{equation}
\mathcal{F}_{\text{DINO}} = \Phi(\mathcal{I}) \in \mathbb{R}^{H' \times W' \times C},
\end{equation}
where $H'$ and $W'$ denote the spatial dimensions of the feature map that are lower than the input dimensions $H$ and $W$, and $C$ represents the feature dimension. Throughout the training of all downstream planning paradigms, the weights of $\Phi$ remain strictly fixed without any gradient updates. This frozen constraint ensures that the perception layer serves as a consistent, domain-invariant interface, preventing the planning head from exploiting visual shortcuts and forcing it to rely on the stable structural priors inherent in the foundation model.

\subsection{Lightweight Feature Adapter}
\label{subsec:lightweight_feature_adaptation}
To bridge the gap between high-dimensional foundation features and planning-specific requirements, we introduce a lightweight feature adapter $\mathcal{A}$ to perform dimensionality reduction and spatial aggregation. Specifically, the adapter first applies a multi-layer perceptron (MLP) to project the raw feature grid $\mathcal{F}_{\text{DINO}} \in \mathbb{R}^{H' \times W' \times C}$ into a compact latent space, producing $\mathcal{F}_{\text{MLP}} \in \mathbb{R}^{H' \times W' \times d}$ with $d < C$, which effectively compresses the feature dimension.
To refine local geometric dependencies and incorporate spatial inductive biases, the feature grid is processed by a CNN-based spatial aggregator. This aggregator follows an upsampling-then-downsampling architecture: it first increases the spatial resolution to facilitate auxiliary perception tasks, such as Bird's-Eye-View (BEV) semantic segmentation, ensuring that the model captures fine-grained structural boundaries. Subsequently, the features are downsampled back to their original dimensions to yield the adapted representation $\mathcal{F}_{\text{ada}}$.
To prepare the representation for the downstream Transformer-based planning decoders, the adapted representation $\mathcal{F}_{\text{ada}}$ is then flattened into a sequence of tokens:
\begin{equation} 
\mathbf{z} = \text{Flatten}(\mathcal{F}_{\text{ada}}) \in \mathbb{R}^{N \times d}, \end{equation}
where $N = H' \times W'$ is the total number of tokens. This sequence constitutes the final adapted DINOv3 representation $\mathbf{z}$, maintaining architectural compatibility with the downstream planning decoders. During training, only the adapter and subsequent planning heads are optimized while the backbone remains strictly frozen. Although the adapter architecture is consistent across all paradigms, each planning head utilizes an independent, non-shared instance of the adapter.

\subsection{Integration with Multi-Paradigm Planners}
\label{Integration_with_Multi_Paradigm_Planners}
To validate the universality and appearance-resilience of the adapted DINOv3 features $\mathbf{z}$, we integrate it into three representative end-to-end planning paradigms: the regression-based LTF \cite{chitta2022transfuser}, the diffusion-based DiffusionDrive \cite{liao2025diffusiondrive}, and the scoring-based GTRS-Dense \cite{li2025generalized}. We derive DINOv3-enhanced variants for these three paradigms, referred to as LTF-DINO, DiffusionDrive-DINO, and GTRS-Dense-DINO, respectively. For each paradigm, the original task-specific backbone is substituted by our frozen foundation-based interface, demonstrating the modularity of the proposed method.

\smallskip
\noindent
\textbf{LTF-DINO}. LTF \cite{chitta2022transfuser} represents a regression-based paradigm that performs imitation learning via direct trajectory regression. The planning head follows a Transformer-based decoder structure that utilizes a sequence of learnable query tokens to interact with the scene context. In LTF-DINO, we adopt the adapted DINOv3 features $\mathbf{z}$ as the scene context. The queries are divided into an ego-query and a set of agent queries. The agent queries are utilized for the DETR-based \cite{DBLP:journals/corr/abs-2005-12872} 3D object detection task to provide multi-task supervision. The ego-query aggregates information from the adapted DINOv3 features $\mathbf{z}$ through cross-attention, which is then decoded by an MLP $f_{\text{reg}}$ to predict the deterministic future trajectory:
\begin{equation}
    \mathcal{T} = \{w_t\}_{t=1}^T = f_{\text{reg}}\!\left(
\mathrm{CrossAttn}(\mathbf{q}_{\text{ego}}, \mathbf{z}),\mathbf{s}
\right),
\end{equation}
where $\mathrm{CrossAttn}$ denotes a cross-attention module that uses 
$\mathbf{q}_{\text{ego}}$ as the query and $\mathbf{z}$ as key and value.

\smallskip
\noindent
\textbf{DiffusionDrive-DINO}. 
DiffusionDrive \cite{liao2025diffusiondrive} models planning as a conditional generative process using a truncated diffusion policy with 20 clustered anchors. Unlike vanilla diffusion models that denoise from pure Gaussian noise, DiffusionDrive initiates the denoising process from an anchored Gaussian distribution centered around prior multi-mode anchors. In DiffusionDrive-DINO, we integrate the adapted DINOv3 features $\mathbf{z}$ as the primary conditional context to guide the denoising steps within an efficient Transformer-based diffusion decoder $f_\text{diff}$. The model learns to recover the trajectory distribution through a truncated schedule, iteratively refining a noisy trajectory $\mathcal{T}_k$ over a limited number of steps:
\begin{equation} 
f_\text{diff}(\mathcal{T}_{k-1} | \mathcal{T}_k, \mathbf{z}, \mathbf{s}),
\end{equation}
where $k$ is the denoising step. 

\smallskip
\noindent
\textbf{GTRS-Dense-DINO}. 
GTRS-Dense-DINO follows the scoring-based paradigm of GTRS-Dense~\cite{li2025generalized}, 
which reformulates trajectory planning as a selection problem over a dense candidate set 
$\mathcal{V}$. The model first employs a trajectory tokenizer to encode each candidate 
trajectory into a latent representation, and a Transformer decoder to model the interaction 
between trajectory candidates and the scene context. In our variant, the adapted DINOv3 
features $\mathbf{z}$ are used as the unified scene context.

Conditioned on the scene feature $\mathbf{z}$ and auxiliary inputs $\mathbf{s}$, the decoder 
outputs are processed by $M$ specialized scoring heads (e.g., NC and DAC), each predicting a 
sub-metric score for a candidate trajectory:
\begin{equation}
    s_{i,j} = g_j(\mathbf{z}, \mathcal{T}_i, \mathbf{s}), 
    \quad j = 1,\dots,M,
\end{equation}
where $g_j$ denotes a scoring head for a sub-metric.

These sub-metrics are aggregated into a unified score via a weighted combination
\begin{equation}
    S_i = \sum_{j=1}^{M} \omega_j s_{i,j},
\end{equation}
where $\omega_j$ denotes the weight associated with the $j$-th criterion.

We denote the entire dense scoring and selection pipeline as a scoring-based planner
\begin{equation}
    \mathcal{T}^* = f_{\text{score}}(\mathbf{z}, \mathbf{s})
    = \arg\max_{\mathcal{T}_i \in \mathcal{V}} S_i,
\end{equation}
which outputs the optimal trajectory by selecting the highest-scoring candidate.

\begin{table*}[t]
\centering
\caption{\textbf{Quantitative results on \benchmarkname, \texttt{navtest} and \texttt{navhard}}. We present comparisons across three planning paradigms. The Method column denotes three model variant methods for each paradigm. For \benchmarkname, in addition to our synthesized out-of-distribution (OOD) images, we report performance on the original images ({\lightgray{Origin}}) as a reference.
}
\label{tab:benchmark_results}
\footnotesize 
\resizebox{\textwidth}{!}{
\begin{tabular}{lc|cccccccccccc|c|c}

\toprule
\multirow{2}{*}{\textbf{Paradigm}} & \multirow{2}{*}{\textbf{Method}} & \multicolumn{12}{c|}{\texttt{\textbf{\benchmarkname}}} & \multicolumn{1}{c|}{\texttt{\textbf{navtest}}} & \multicolumn{1}{c}{\texttt{\textbf{navhard}}} \\ \cmidrule{3-16}
 & & \textbf{Appearance} & \textbf{NC}$\uparrow$ & \textbf{DAC}$\uparrow$ & \textbf{DDC}$\uparrow$ & \textbf{TLC}$\uparrow$ & \textbf{EP}$\uparrow$ & \textbf{TTC}$\uparrow$ & \textbf{LK}$\uparrow$ & \textbf{HC}$\uparrow$ & \textbf{EC}$\uparrow$ & \textbf{EPDMS}$\uparrow$ & \textbf{Drop Rate}$\downarrow$ & \textbf{EPDMS}$\uparrow$ & \textbf{EPDMS}$\uparrow$ \\

\midrule
\multicolumn{16}{c}{\textit{Regression-based Planner}} \\ 
\midrule

\multirow{6}{*}{LTF \cite{chitta2022transfuser}} & \multirow{2}{*}{Base} & \lightgray{Origin} & \lightgray{97.8} & \lightgray{92.4} & \lightgray{99.3} & \lightgray{99.6} & \lightgray{87.7} & \lightgray{96.8} & \lightgray{95.6} & \lightgray{97.4} & \lightgray{80.3} & \lightgray{84.5} & \multirow{2}{*}{10.1\%} & \multirow{2}{*}{84.7} & \multirow{2}{*}{25.2} \\ 
 & & OOD & 94.7 & 87.5 & 97.5 & 99.5 & 86.2 & 93.5 & 92.0 & 97.3 & 78.3 & 76.0 & & & \\
\cmidrule{2-16}
 & \multirow{2}{*}{DR} & \lightgray{Origin} & \lightgray{98.1} & \lightgray{92.9} & \lightgray{99.5} & \lightgray{99.6} & \lightgray{87.1} & \lightgray{97.3} & \lightgray{95.6} & \lightgray{97.4} & \lightgray{78.5} & \lightgray{85.2} & \multirow{2}{*}{3.8\%} & \multirow{2}{*}{86.0} & \multirow{2}{*}{30.2} \\
 & & OOD & 97.2 & 90.8 & \textbf{99.0} & \textbf{99.7} & 85.8 & 96.2 & 94.1 & 97.3 & 79.3 & 81.9 & & & \\
\cmidrule{2-16}
 & \multirow{2}{*}{DINO (Ours)} & \lightgray{Origin} & \lightgray{97.8} & \lightgray{94.0} & \lightgray{99.2} & \lightgray{99.7} & \lightgray{87.4} & \lightgray{96.3} & \lightgray{96.0} & \lightgray{97.2} & \lightgray{81.2} & \lightgray{85.7} & \multirow{2}{*}{\textbf{1.0\%}} & \multirow{2}{*}{\textbf{86.6}} & \multirow{2}{*}{\textbf{30.4}} \\
 & & OOD & \textbf{97.3} & \textbf{93.6} & \textbf{99.0} & \textbf{99.7} & \textbf{87.4} & \textbf{96.3} & \textbf{94.6} & 97.3 & \textbf{81.2} & \textbf{84.8} & & & \\

\midrule
\multicolumn{16}{c}{\textit{Diffusion-based Planner}} \\
\midrule

\multirow{6}{*}{DiffusionDrive \cite{liao2025diffusiondrive}} & \multirow{2}{*}{Base} & \lightgray{Origin} & \lightgray{98.4} & \lightgray{94.4} & \lightgray{99.5} & \lightgray{99.7} & \lightgray{87.6} & \lightgray{97.6} & \lightgray{96.4} & \lightgray{97.4} & \lightgray{82.1} & \lightgray{87.3} & \multirow{2}{*}{7.9\%} & \multirow{2}{*}{86.8} & \multirow{2}{*}{27.1} \\
 & & OOD & 96.6 & 90.0 & 98.2 & \textbf{99.6} & 86.1 & 95.5 & 92.7 & \textbf{97.3} & 80.1 & 80.4 & & & \\
\cmidrule{2-16}
 & \multirow{2}{*}{DR} & \lightgray{Origin} & \lightgray{98.2} & \lightgray{94.3} & \lightgray{99.3} & \lightgray{99.6} & \lightgray{87.7} & \lightgray{97.5} & \lightgray{96.0} & \lightgray{97.4} & \lightgray{82.1} & \lightgray{86.6} & \multirow{2}{*}{5.3\%} & \multirow{2}{*}{\textbf{87.6}} & \multirow{2}{*}{29.5} \\
 & & OOD & 96.9 & 90.9 & 98.8 & \textbf{99.6} & 87.0 & 95.9 & 93.7 & \textbf{97.3} & \textbf{81.0} & 82.0 & & & \\
\cmidrule{2-16}
 & \multirow{2}{*}{DINO (Ours)} & \lightgray{Origin} & \lightgray{98.3} & \lightgray{95.0} & \lightgray{99.7} & \lightgray{99.6} & \lightgray{87.2} & \lightgray{97.6} & \lightgray{96.3} & \lightgray{97.4} & \lightgray{79.4} & \lightgray{87.3} & \multirow{2}{*}{\textbf{0.7\%}} & \multirow{2}{*}{\textbf{87.6}} & \multirow{2}{*}{\textbf{33.1}} \\
 & & OOD & \textbf{97.7} & \textbf{95.0} & \textbf{99.4} & \textbf{99.6} & \textbf{87.4} & \textbf{96.8} & \textbf{96.0} & \textbf{97.3} & 79.5 & \textbf{86.7} & & &  \\

\midrule
\multicolumn{16}{c}{\textit{Scoring-based Planner}} \\
\midrule

\multirow{6}{*}{GTRS-Dense \cite{li2025generalized}} & \multirow{2}{*}{Base} & \lightgray{Origin} & \lightgray{99.0} & \lightgray{98.1} & \lightgray{99.4} & \lightgray{99.8} & \lightgray{81.8} & \lightgray{98.9} & \lightgray{93.6} & \lightgray{97.1} & \lightgray{44.8} & \lightgray{87.9} & \multirow{2}{*}{5.8\%} & \multirow{2}{*}{84.8} & \multirow{2}{*}{45.4} \\
 & & OOD & 97.7 & 96.4 & 99.0 & \textbf{99.8} & 75.9 & 97.7 & 92.0 & 96.5 & 38.7 & 82.9 & & & \\
\cmidrule{2-16}
 & \multirow{2}{*}{DR} & \lightgray{Origin} & \lightgray{98.7} & \lightgray{98.1} & \lightgray{99.5} & \lightgray{99.9} & \lightgray{81.5} & \lightgray{98.6} & \lightgray{95.1} & \lightgray{96.9} & \lightgray{44.4} & \lightgray{87.9} & \multirow{2}{*}{3.1\%} & \multirow{2}{*}{\textbf{86.0}} & \multirow{2}{*}{\textbf{46.8}} \\
 & & OOD & \textbf{98.3} & 97.3 & \textbf{99.1} & \textbf{99.8} & 78.8 & \textbf{98.2} & \textbf{93.4} & 96.8 & 41.8 & 85.2 & & & \\
\cmidrule{2-16}
 & \multirow{2}{*}{DINO (Ours)} & \lightgray{Origin} & \lightgray{98.5} & \lightgray{98.5} & \lightgray{99.5} & \lightgray{99.7} & \lightgray{82.2} & \lightgray{97.9} & \lightgray{93.6} & \lightgray{97.1} & \lightgray{49.8} & \lightgray{87.8} & \multirow{2}{*}{\textbf{0.7\%}} & \multirow{2}{*}{85.8} & \multirow{2}{*}{46.7} \\
 & & OOD & \textbf{98.3} & \textbf{98.2} & \textbf{99.1} & 99.7 & \textbf{82.4} & 97.7 & 92.7 & \textbf{97.1} & \textbf{52.1} & \textbf{87.2} & & & \\
 
\bottomrule
\end{tabular}
}
\end{table*}

\section{Experiments}
\label{sec:experiments}

\subsection{Experimental Setup}

\smallskip
\noindent
\textbf{Benchmarks:} We assess the generalization of planning capabilities across both out-of-distribution styles and standard driving scenarios using the following benchmarks:
\textbf{1) \benchmarkname:} 
To facilitate a rigorous comparison between structural robustness and data-driven augmentation, \benchmarkname is partitioned into a 40\% Augmentation Support Set and a 60\% Evaluation Set based on spatial layouts. Within these sets, the 10 visual styles are categorized into two disjoint groups: 5 ``Seen Styles'' utilized \textit{exclusively} for training the domain randomization (DR) baseline, and 5 ``Unseen Styles'' reserved for zero-shot testing. All final evaluations are performed strictly on the ``Unseen Styles'' subset of the Evaluation Set to ensure a fair OOD assessment. \textbf{2) \texttt{navtest} from NAVSIMv1 \cite{Dauner2024NEURIPS}:} 
NAVSIMv1 offers the \texttt{navtest} set, which is a one-stage evaluation benchmark containing 12,146 real-world scenarios. 
\textbf{3) \texttt{navhard} from NAVSIMv2 \cite{Cao2025CORL}:}
NAVSIMv2 offers a more challenging \texttt{navhard} set, comprising 244 real-world scenarios and 4,164 synthetic scenarios through 3DGS \cite{kerbl20233d} for pseudo-closed-loop evaluation with reactive traffic participants governed by the Intelligent Driver Model (IDM) \cite{treiber2000congested}. \par

\smallskip
\noindent
\textbf{Evaluation Metrics}: For all benchmarks, we employ EPDMS as our primary metric as described in \cref{sec:evaluation_metrics}.

\smallskip 
\noindent 
\textbf{Comparison Methods:} To demonstrate the advantages of the frozen foundation’s robustness over conventional robustness-oriented methods, we define three comparative variants for each planning paradigm: 
     \textbf{1) Base}: The original planning model utilizing a supervised backbone trained exclusively on the standard \texttt{navtrain} split. 
    \textbf{2) DR}: A Domain Randomization baseline where the original supervised model is co-trained on the \texttt{navtrain} split and the \benchmarkname Augmentation Support Set using the ``Seen Styles'' for augmentation.
    \textbf{3) DINO}: Our proposed integration of the frozen DINOv3 backbone, trained solely on the standard \texttt{navtrain} split without any exposure to the synthetic visual styles in \benchmarkname, with the same training setup as the Base variant.

\subsection{Implementation Details}
We evaluate LTF and GTRS-Dense using their official released checkpoints without modification. For DiffusionDrive, we implement and train an image-only variant, referred to as the latent version, to facilitate a direct comparison within our visual-centric framework. For DINO-based variants (LTF-DINO, DiffusionDrive-DINO, and GTRS-Dense-DINO), we train only the feature adapters and planning heads on top of a frozen DINOv3 backbone. Training is conducted on a server equipped with 8 NVIDIA RTX 3090 GPUs. Detailed network architecture and training hyperparameters are provided in the supplementary material.

\subsection{Quantitative Comparison}

\smallskip
\noindent
\textbf{Results on \benchmarkname.} As shown in \cref{tab:benchmark_results}, we compare the performance of all three planning paradigms across original appearances and visual perturbations. The original supervised Base models exhibit a catastrophic performance collapse when transitioning from the Origin to OOD appearance scenarios, highlighting their sensitivity to domain shifts. Since \benchmarkname utilizes pixel-aligned style transfer, the underlying semantic and geometric properties are largely preserved relative to the original frames. Consequently, the failure of baseline models directly reveals their over-reliance on texture-centric features for decision-making. Notably, while the DR baseline, which is explicitly fine-tuned on ``Seen Styles'', shows improved resilience compared to the Base model, its performance remains significantly inferior to DINO-based variants when facing ``Unseen Styles.'' This performance gap suggests that domain randomization primarily encourages the model to adapt to specific localized style distributions rather than abstracting consistent semantic invariants. In contrast, the models equipped with the frozen DINOv3 backbone demonstrate exceptional stability, with only marginal performance degradation, despite not being trained on augmented styles. We attribute this robustness to DINOv3’s capacity for extracting semantic priors that are invariant to visual appearance, which we will discuss in \cref{sec:analysis}.

\smallskip
\noindent
\textbf{Results on \texttt{navtest} and \texttt{navhard}.} 
As shown in \cref{tab:benchmark_results}, the integration of the frozen DINOv3 backbone consistently outperforms the original supervised backbones across all three planning paradigms on the official NAVSIM benchmarks. This indicates that our method not only provides a robust solution for OOD appearance shifts but also enhances overall planning performance in standard driving scenarios.

\subsection{Qualitative Comparison}
\label{sec:qualitative_comparison}

\definecolor{expertgreen}{RGB}{0, 255, 0}
\definecolor{predred}{RGB}{231, 112, 97}

\begin{figure*}[t]
    \centering
    \includegraphics[width=\textwidth]{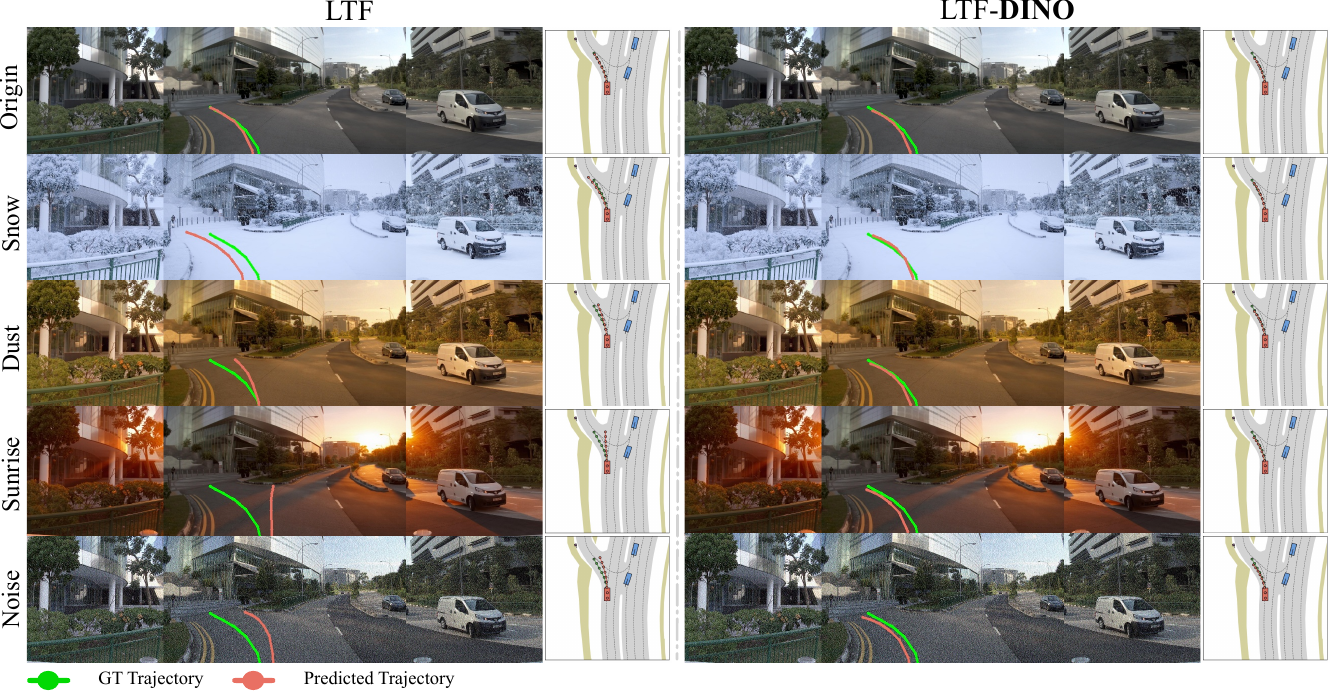}
    \caption{\textbf{Qualitative comparison of the planning results.} We visualize the planning performance of the LTF baseline (left) and LTF-DINO (right) across diverse domains. While the baseline exhibits hazardous trajectories when encountering visual domain shifts, LTF-DINO consistently produces robust, consistent, and human-aligned paths despite severe environmental perturbations.}
    \label{fig:qualitative_results}
\end{figure*}

To provide deeper insights into the generalization capabilities of different backbones, we visualize the predicted trajectories under diverse visual domains. \cref{fig:qualitative_results} presents a comparative visualization between the baseline LTF and LTF-DINO across five distinct scenarios.
As observed in the Origin scenario, both the baseline LTF and LTF-DINO deliver satisfactory performance when operating within the training domain. Their predicted trajectories closely align with the human trajectory, indicating that both backbones can effectively learn driving policies under standard, clean conditions.\par
Nevertheless, a significant disparity in performance arises in out-of-distribution scenarios. The baseline LTF exhibits severe performance degradation under style perturbations. For instance, in the Heavy Snow scenario, the baseline's trajectory deviates significantly from the road, leading to a collision with the curb. Similar deviations are observed in other settings. On the contrary, LTF-DINO demonstrates remarkable resilience to these visual corruptions. Across all the challenging scenarios, the trajectories predicted by LTF-DINO remain stable and smooth, and consistently stay within the safe lane boundaries, mirroring its performance in the Origin domain.

\subsection{Analysis}
\label{sec:analysis}

To gain a deeper understanding of the inherent robustness and decision-making logic of DINOv3-based representations, we perform a multi-dimensional qualitative analysis. Our investigation spans three distinct perspectives: feature-level semantic stability, latent spatial consistency, and internal decision-making analysis through attention maps.

\begin{figure}[t]
    \centering
    \includegraphics[width=\linewidth]{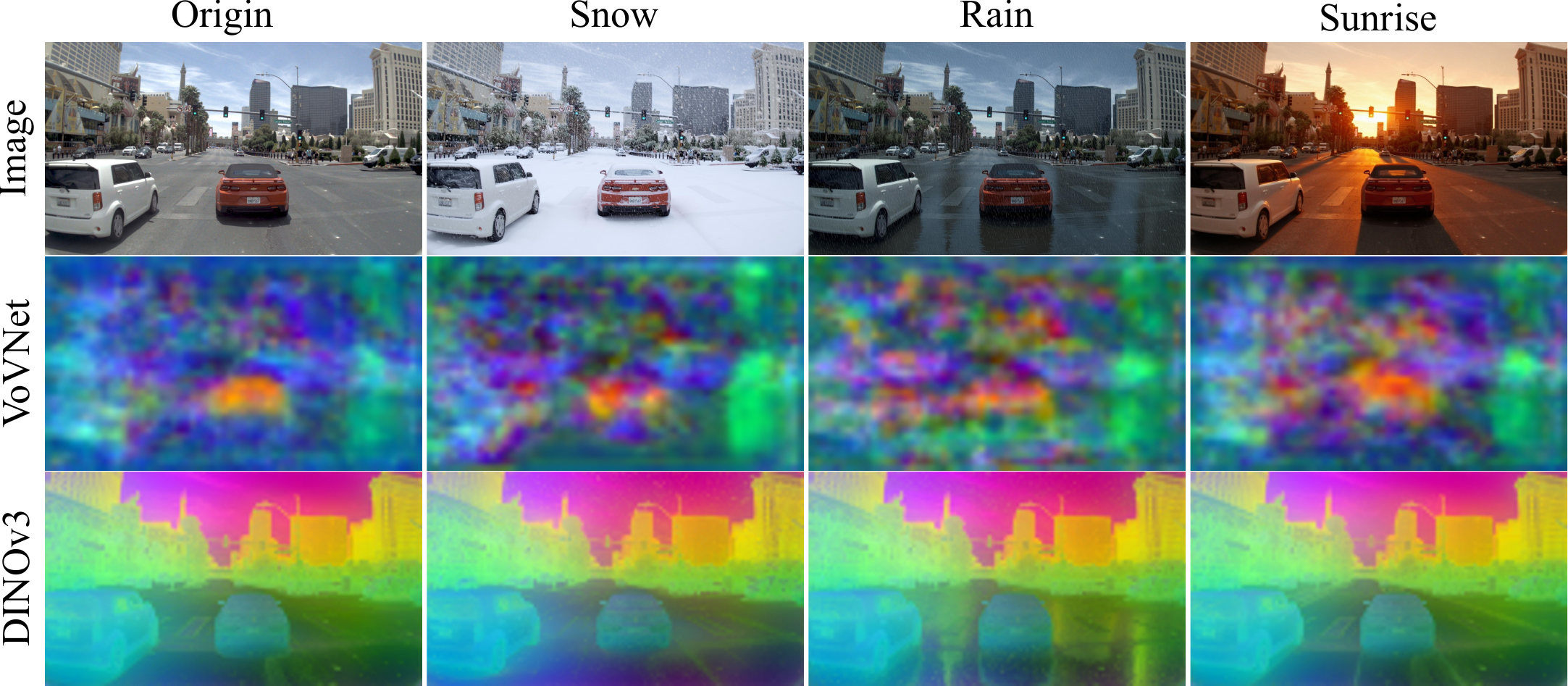}
    \caption{\textbf{Feature visualization.} For each appearance, we show the input image, VoVNet's PCA feature map, and DINOv3's PCA feature map. While DINOv3 extracts appearance-invariant semantic structures across style shifts, the VoVNet backbone exhibits large variations.}
    \label{fig:appearance_pca}
\end{figure}

\smallskip
\noindent
\textbf{Comparison of Representation Robustness.} We perform PCA on the feature maps extracted by the frozen DINOv3-H+ backbone and the VoVNet \cite{lee2019energy} backbone employed in GTRS-Dense \cite{li2025generalized} on the same driving scene under multiple visual appearances. As shown in \cref{fig:appearance_pca}, the PCA feature maps generated by the DINOv3 backbone remain consistent, effectively preserving the core spatial layout and scene geometry with minimal distortion. In contrast, the VoVNet backbone fails to maintain structural coherence under style shifts and the feature maps become fragmented and noisy. This comparison highlights the superior appearance-invariance of DINOv3, which is essential for robust zero-shot generalization.

\begin{figure}[t]
    \centering
    \includegraphics[width=0.49\textwidth]{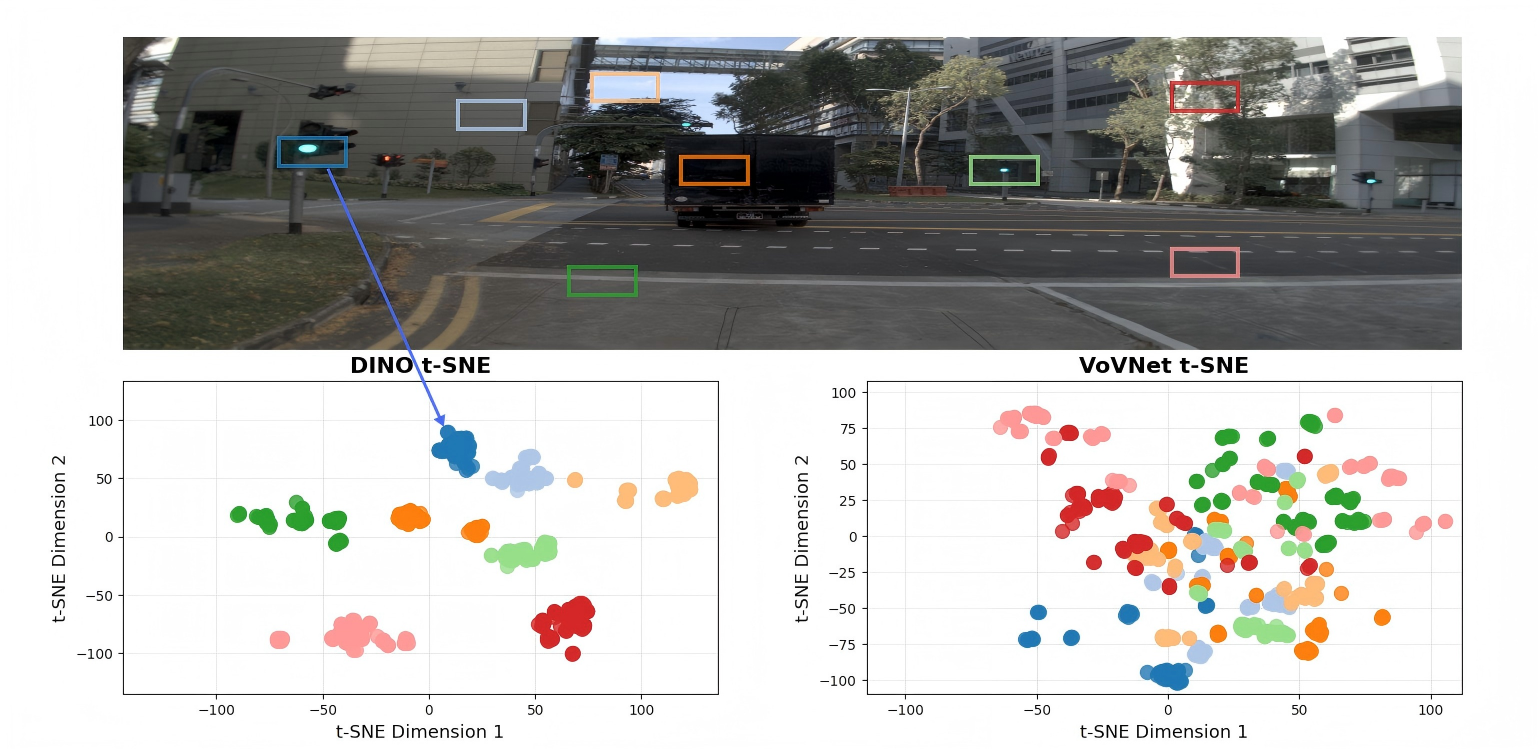}
    \caption{\textbf{t-SNE visualization of spatial token consistency.} We compare the feature tokens from DINOv3 and VoVNet for a single scene across all visual domains. Points are color-coded by their original spatial coordinates.}
    \label{fig:t-SNE}
\end{figure}

\smallskip
\noindent
\textbf{Spatial Consistency Analysis via t-SNE}. To further quantify the appearance-invariance of the extracted features, we perform the t-SNE visualization using spatial tokens from a single scene across all visual domains. For visual clarity, we selectively plot representative regions where each point represents a token within a $3 \times 3$ spatial neighborhood, color-coded by its coordinate position. \cref{fig:t-SNE} shows that DINOv3 tokens from the same spatial location form tight clusters across diverse appearance shifts, demonstrating that the representation remains invariant to visual perturbations. In contrast, the VoVNet tokens exhibit significant dispersion, indicating that its latent space is highly sensitive to domain shifts.

\begin{figure}[t]
    \centering
    \includegraphics[width=\linewidth]{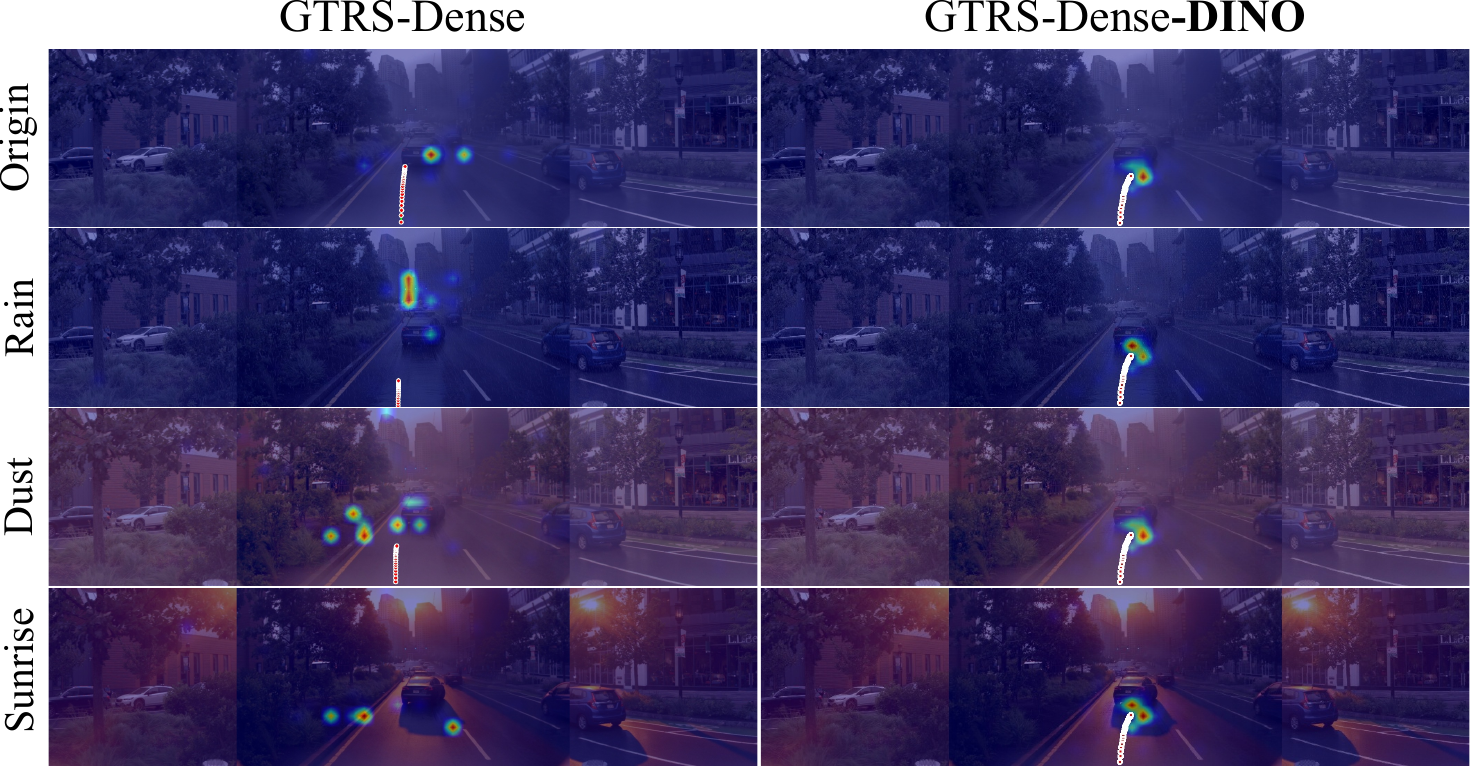}
    \caption{\textbf{Qualitative visualization of attention maps and planned trajectories across visual domains.} We compare the last-layer attention maps and the planned trajectory of GTRS-Dense and GTRS-Dense-DINO within the same scenario subjected to different visual domains. GTRS-Dense predicts a near-zero velocity trajectory at the bottom row, hence invisible.}
    \label{fig:attention_map}
\end{figure}

\smallskip
\noindent
\textbf{Comparison of Planning Focus via Attention Maps}. To further investigate the internal decision-making process, we visualize the attention maps from the final layer of the scoring-based paradigm GTRS-Dense, focusing on the trajectory with the highest predicted score across various appearance shifts. As illustrated in \cref{fig:attention_map}, the baseline GTRS-Dense effectively focuses on the preceding vehicle in the Origin domain, yielding a safe and standard trajectory. However, when subjected to the targeted visual stress tests in \benchmarkname, the baseline’s attention becomes scattered and unstable, failing to maintain focus on critical scene features. This perceptual breakdown results in unreliable and hazardous trajectories that deviate from the drivable area. In contrast, GTRS-Dense-DINO maintains a precise and consistent focus on the lead vehicle regardless of severe appearance shifts. By leveraging appearance-invariant semantic priors, GTRS-Dense-DINO ensures that the planning logic remains grounded in the underlying scene structure, resulting in highly stable and expert-aligned trajectories.

\subsection{Ablation}

\begin{table}[t]
\centering
\caption{\textbf{Ablation on training strategies using DINOv3-B}. We compare End-to-End (E2E) fine-tuning and the Frozen Backbone approach, reporting EPDMS scores on the standard \texttt{navtest} and the Origin/OOD scenarios of \benchmarkname.}
\label{tab:ablation_frozen}
\resizebox{\columnwidth}{!}{
\begin{tabular}{lccc}
\toprule
\multirow{2.5}{*}{Strategy} & \multirow{2.5}{*}{\shortstack{\textbf{\texttt{navtest}} \\ (EPDMS $\uparrow$)}} & \multicolumn{2}{c}{\textbf{\benchmarkname} (EPDMS $\uparrow$)} \\
\cmidrule(lr){3-4}
& & Origin & OOD \\
\midrule
E2E & 86.8 & 86.4 & 79.9 \scriptsize{\color{red}{(-7.5\%)}} \\
Frozen Backbone & 86.4 & 86.4 & 84.5 \scriptsize{\color{blue}{(-2.2\%)}} \\
\bottomrule
\end{tabular}
}
\end{table}

\begin{table}[t]
\centering
\caption{\textbf{Ablation on adapter configurations} for LTF-DINO. We evaluate the impact of MLP layer (L) depth and the integration of a CNN-based spatial aggregator on \texttt{navtest}.}
\label{tab:ablation_adapter}
\resizebox{\columnwidth}{!}{ 
\begin{tabular}{@{}l|cccc@{}}
\toprule
\textbf{Config} & 2L & 4L & 4L+CNN & 8L+CNN \\ \midrule
\textbf{navtest} (EPDMS$\uparrow$) & 81.4 & 84.4 & \textbf{86.6} & 86.4 \\ \bottomrule
\end{tabular}
} 
\end{table}

\smallskip
\noindent
\textbf{The Necessity of Frozen Foundation Backbones.} To investigate the impact of backbone adaptation on planning performance and robustness, we compare two training strategies for the LTF-DINO model: (1) E2E fine-tuning, where the backbone is updated alongside the planning head, and (2) Frozen Backbone, where the backbone parameters remain fixed. For these experiments, we utilize DINOv3-B \cite{simeoni2025dinov3} as the backbone to ensure computational efficiency.
As shown in \cref{tab:ablation_frozen}, the E2E fine-tuned model achieves marginally superior performance on the \texttt{navtest} compared to its frozen version. This minor gain suggests that fine-tuning allows the backbone to over-specialize in task-specific textural cues and dataset-specific biases. However, this adaptation comes at a significant cost to robustness. The E2E model suffers a substantial performance degradation on \benchmarkname, whereas the frozen version exhibits greater robustness.

\smallskip
\noindent
\textbf{Impact of Feature Adapter Configurations.} We investigate the influence of adapter depth and architecture on LTF-DINO using the \texttt{navtest} set. As shown in \cref{tab:ablation_adapter}, increasing the depth of MLP from 2 to 4 layers yields a significant 3.0\% improvement in EPDMS, indicating that sufficient capacity is required to project the foundation features into the planning space. Notably, integrating a CNN-based spatial aggregator with the 4-layer MLP achieves the peak performance. However, expanding the adapter to 8 layers results in a slight decline, suggesting that deeper adapters do not yield additional gains.

\section{Conclusion}  
\label{sec:conclusion}
In this work, we isolated the specific impact of visual appearance on E2E-AD through \benchmarkname, a high-fidelity generative stress-test benchmark, revealing that standard models are highly fragile to such perturbations despite constant scene geometry. To address this, we validated the use of a frozen DINOv3 backbone as a universal, appearance-invariant perception interface. Our results confirm that this plug-and-play solution achieves exceptional zero-shot generalization across different planning paradigms, including regression, diffusion, and scoring-based models, without any domain-specific training. 
Since our method provides an effective solution to address appearance OOD, future research can pivot its focus toward investigating structural OOD challenges arising from diverse and complex scenes.

\bibliographystyle{plainnat}
\bibliography{references}

\clearpage
\appendix

\begin{table*}[t!]
\centering
\caption{\textbf{Detailed Prompts for \benchmarkname OOD Conditions.}}
\label{tab:prompts}
\begin{tabularx}{\textwidth}{l|X}
\toprule
\textbf{Condition} & \textbf{Specific Text Prompt} \\
\midrule
\rowcolor{gray!10} \multicolumn{2}{c}{\textit{Weather}} \\
Light Dust & The image should show light dust or airborne particles, with light scattering, overall yellowish-brown tones, and slightly reduced visibility. \\
Heavy Snow & Transform the image into a heavy snow scene with dense snowflakes, reduced visibility, and ground covered with snow. \\
Heavy Rain & Transform the image into a heavy rain scene with dense raindrops, slippery ground with visible water splashes, and slightly blurred distant views. \\
\midrule
\rowcolor{gray!10} \multicolumn{2}{c}{\textit{Lighting}} \\
Dawn Sunrise & The image should show dawn or sunrise lighting, with low-angle warm-toned sunlight creating long and soft shadows. \\
Dusk Sunset & The image should show dusk or sunset lighting, with low-angle warm-toned light, long shadows, and overall orange-red or golden tones. \\
Dappled Light & The image should show complex, uneven dappled light and shadow, simulating light passing through leaves or complex structures projecting onto the ground and objects. \\
\midrule
\rowcolor{gray!10} \multicolumn{2}{c}{\textit{Style}} \\
Vintage Photo & Transform the image into a vintage photo style, with faded colors, yellow or sepia tones, possibly with grain and slight scratches. \\
CARLA Toy-like & Transform the image into a game engine rendered asset style, with objects looking like digital models or toys, slightly saturated colors, clear edges, and non-photorealistic appearance. \\
\midrule
\rowcolor{gray!10} \multicolumn{2}{c}{\textit{Effects}} \\
Motion Blur & Add obvious motion blur effects to the image, simulating camera movement during fast shooting or fast-moving objects in the frame. \\
Digital Noise & Add a large amount of digital noise to the image, especially in dark areas, simulating low light or high ISO settings. \\
\bottomrule
\end{tabularx}
\end{table*}

\subsection{Generative Prompts for \benchmarkname}
We detail the specific text prompts used in our generative pipeline. As shown in \cref{tab:prompts}, these prompts cover four major domains: Weather, Lighting, Style, and Effects, providing a comprehensive stress test for appearance robustness.

\subsection{Detailed Architectures of Planning Paradigms}
\begin{figure*}[t!]
    \centering
    \includegraphics[width=\textwidth]{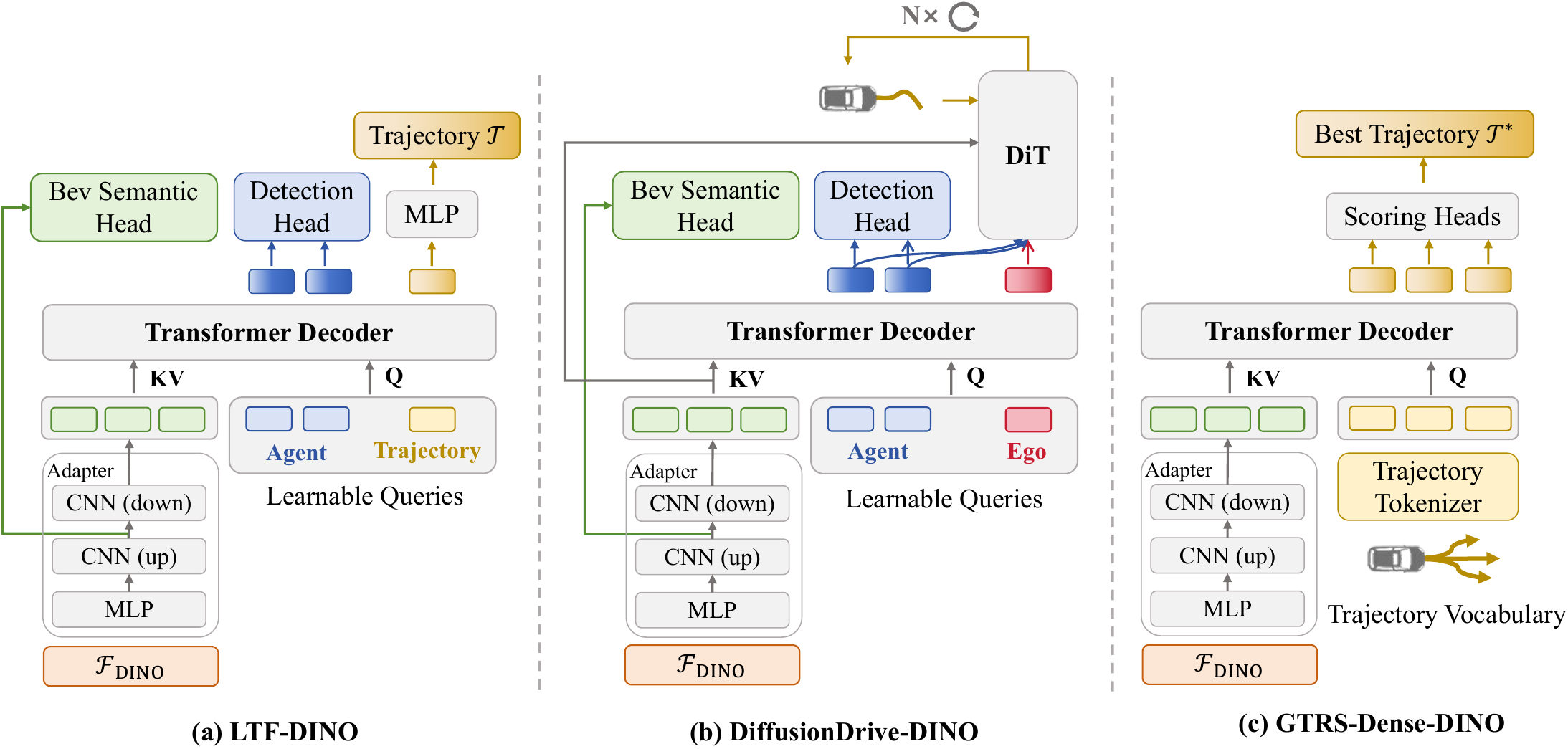}
    \caption{\textbf{Detailed Architectures of Planning Paradigms.} We illustrate the detailed network architectures for (a) LTF-DINO, (b) DiffusionDrive-DINO, and (c) GTRS-Dense-DINO.}
    \label{fig:detailed_architectures}
\end{figure*}

While the main manuscript outlines the high-level integration of the proposed Constant Eye interface across different planning paradigms, here we provide the detailed internal network architectures and data flows for each DINO-enhanced head. As illustrated in \cref{fig:detailed_architectures}, in all three paradigms, the robust semantic features $\mathcal{F}_{\text{DINO}}$ from the frozen backbone are first processed by a paradigm-specific adapter to obtain the adapted DINOv3 features $\mathbf{z}$. This adapted representation $\mathbf{z}$ is then injected into the respective planning heads as the main scene context.

\smallskip
\noindent
\textbf{LTF-DINO.} As shown in \cref{fig:detailed_architectures} (a), the regression head adopts a Transformer decoder architecture. The adapted representation $\mathbf{z}$ serves as the robust keys and values for the attention mechanism. Learnable Agent Queries and a Trajectory Query serve as queries to extract relevant semantic information via cross-attention layers. The updated agent queries are subsequently decoded by an auxiliary detection head, while the updated trajectory query is processed by an MLP regression head to directly output the future waypoints.

\smallskip
\noindent
\textbf{DiffusionDrive-DINO.} In the diffusion-based paradigm (\cref{fig:detailed_architectures} (b)), the planning task is formulated as a conditional denoising process. The adapted feature $\mathbf{z}$ serves as the robust scene context for a Transformer decoder, where learnable Agent Tokens and Ego Tokens act as queries. Similar to the regression-based approach, the refined agent tokens are supervised by an auxiliary detection head. Meanwhile, the refined ego tokens, the adapted feature $\mathbf{z}$, and the agent tokens collectively serve as the condition for a Conditional Diffusion-Transformer (DiT). This DiT block iteratively refines the trajectory distribution through a designated number of denoising steps to generate the final trajectory.

\smallskip
\noindent
\textbf{GTRS-DINO.} For the scoring-based GTRS (\cref{fig:detailed_architectures} (c)), $\mathbf{z}$ is utilized to provide the necessary semantic context for evaluating static trajectory candidates. A predefined trajectory vocabulary is first processed by a trajectory tokenizer to generate trajectory tokens. These tokens then serve as queries in a Transformer decoder to extract specific semantic features from $\mathbf{z}$ that correspond to each candidate trajectory. The resulting trajectory token features are passed through scoring heads to predict confidence scores for all candidates. Finally, an $\operatorname{argmax}$ operation is applied to select the trajectory with the highest score as the optimal trajectory $\mathcal{T}^*$.

\subsection{Additional Implementation Details}
As summarized in \cref{tab:implementation_details}, the DINO interface is integrated with paradigm-specific configurations. We utilize official checkpoints for LTF \cite{chitta2022transfuser} and GTRS-Dense \cite{li2025generalized}, while training DiffusionDrive \cite{liao2025diffusiondrive} from scratch using training configurations identical to those for DiffusionDrive-DINO. LTF-DINO and DiffusionDrive-DINO are trained for 100 epochs with auxiliary detection and segmentation tasks to predict trajectories for a 4-second future horizon at an operating frequency of 2Hz. In contrast, GTRS-Dense-DINO is trained for 50 epochs without auxiliary supervision at an operating frequency of 10Hz while maintaining the same prediction window. All DINO modules are optimized via AdamW with tailored learning rates and schedules to ensure stable convergence.

\vspace{0.5em} 
\noindent\begin{minipage}{\columnwidth}
\centering
\captionof{table}{\textbf{Implementation details and model configurations.} CKPT, Det., and Seg. denote checkpoint, detection, and segmentation, respectively.}
\label{tab:implementation_details}
\footnotesize
\resizebox{\columnwidth}{!}{
\begin{tabular}{l|ccc}
\toprule
\textbf{Configuration} & \textbf{LTF} & \textbf{DiffusionDrive} & \textbf{GTRS-Dense} \\
\midrule
Model Source & Official CKPT & From scratch & Official CKPT \\
Backbone & ResNet-34 \cite{he2016deep} & ResNet-34 \cite{he2016deep} & V2-99 \cite{lee2019energy} \\
Sensors & 3 $\times$ Cam & 3 $\times$ Cam & 3 $\times$ Cam \\
Resolution & 1024 $\times$ 256 & 1024 $\times$ 256 & 2048 $\times$ 512 \\
\midrule
\rowcolor{gray!10} \multicolumn{4}{l}{\textit{DINO Interface Settings}} \\
\midrule
Backbone & DINOv3-H+ \cite{simeoni2025dinov3} & DINOv3-H+ \cite{simeoni2025dinov3} & DINOv3-H+ \cite{simeoni2025dinov3} \\
Resolution & 1024 $\times$ 256 & 1024 $\times$ 256 & 1024 $\times$ 256 \\
Horizon & 4s & 4s & 4s\\
Frequency & 2Hz & 2Hz & 10Hz \\
Aux.Tasks & Det. \& Seg. & Det. \& Seg. & None \\
Epochs & 100 & 100 & 50 \\
Batch Size & 256 & 512 & 256 \\
Learning Rate & $3\times 10^{-4}$ & $6\times 10^{-4}$ & $1\times 10^{-4}$ \\
LR Schedule & Constant & Cosine Decay & Cosine Decay \\
Optimizer & AdamW & AdamW & AdamW \\
\bottomrule
\end{tabular}
}
\end{minipage}
\vspace{0.5em}

\subsection{Detailed Performance Metrics on navtest and navhard}
We report the comprehensive breakdown of the EPDMS sub-metrics for both \texttt{navtest} and \texttt{navhard} benchmarks in \cref{tab:navtest_detailed} and \cref{tab:navhard_detailed}. These detailed results offer a complete view of the performance across all individual evaluation dimensions that could not be fully included in the main manuscript due to space constraints.
\begin{table*}[t!]
\centering
\caption{\textbf{Detailed Performance Breakdown on \texttt{navtest}.} We present the comprehensive sub-metric results for the three planning paradigms. The Method column denotes model variants: Base (standard backbones), DR (Domain Randomization), and DINO (our Constant Eye interface). All results are reported on the \texttt{navtest} benchmark.}
\label{tab:navtest_detailed}
\footnotesize 
\resizebox{\textwidth}{!}{
\begin{tabular}{lc|ccccccccc|c}
\toprule
\textbf{Paradigm} & \textbf{Method} & \textbf{NC}$\uparrow$ & \textbf{DAC}$\uparrow$ & \textbf{DDC}$\uparrow$ & \textbf{TLC}$\uparrow$ & \textbf{EP}$\uparrow$ & \textbf{TTC}$\uparrow$ & \textbf{LK}$\uparrow$ & \textbf{HC}$\uparrow$ & \textbf{EC}$\uparrow$ & \textbf{EPDMS}$\uparrow$ \\ 

\midrule
\multicolumn{12}{c}{\textit{Regression-based Planner}} \\ 
\midrule
\multirow{3}{*}{LTF \cite{chitta2022transfuser}} & Base & 97.8 & 92.8 & 99.4 & \textbf{99.8} & 87.3 & 97.0 & 96.6 & \textbf{98.4} & 87.2 & 84.7 \\
 & DR & \textbf{98.5} & 93.9 & \textbf{99.5} & \textbf{99.8} & 86.9 & \textbf{97.8} & \textbf{97.1} & 98.2 & 83.9 & 86.0 \\
 & DINO (Ours) & 97.9 & \textbf{94.9} & 99.4 & \textbf{99.8} & \textbf{87.4} & 96.9 & 96.8 & 98.3 & \textbf{87.3} & \textbf{86.6} \\

\midrule
\multicolumn{12}{c}{\textit{Diffusion-based Planner}} \\
\midrule
\multirow{3}{*}{DiffusionDrive \cite{liao2025diffusiondrive}} & Base & 98.2 & 94.7 & 99.4 & \textbf{99.8} & 87.3 & 97.3 & 96.8 & \textbf{98.3} & \textbf{87.5} & 86.8 \\
 & DR & \textbf{98.4} & 95.3 & 99.4 & \textbf{99.8} & \textbf{87.6} & \textbf{97.5} & 97.1 & \textbf{98.3} & 87.2 & \textbf{87.6} \\
 & DINO (Ours) & 98.1 & \textbf{95.6} & \textbf{99.5} & \textbf{99.8} & 87.3 & 97.4 & \textbf{97.3} & \textbf{98.3} & 86.8 & \textbf{87.6} \\

\midrule
\multicolumn{12}{c}{\textit{Scoring-based Planner}} \\
\midrule
\multirow{3}{*}{GTRS-Dense \cite{li2025generalized}} & Base & \textbf{99.1} & 97.9 & 99.5 & 99.9 & 81.9 & \textbf{98.8} & 94.7 & \textbf{98.2} & 48.7 & 84.8 \\
 & DR & \textbf{99.1} & \textbf{98.7} & \textbf{99.6} & \textbf{100.0} & 81.8 & \textbf{98.8} & \textbf{95.8} & 98.1 & 51.3 & \textbf{86.0} \\
 & DINO (Ours) & 98.6 & 98.6 & \textbf{99.6} & 99.9 & \textbf{82.7} & 98.3 & 94.6 & \textbf{98.2} & \textbf{55.3} & 85.8 \\
 
\bottomrule
\end{tabular}
}
\end{table*}

\begin{table*}[t!]
\centering
\caption{\textbf{Detailed Performance Breakdown on \texttt{navhard}.} We report sub-metrics across two evaluation stages: \textbf{S1} and \textbf{S2} (3DGS rendered) on \texttt{navhard}. The Score column represents the specific performance for each stage, while EPDMS denotes the overall benchmark result.}
\label{tab:navhard_detailed}
\footnotesize 
\resizebox{\textwidth}{!}{
\begin{tabular}{lc|c|ccccccccc|c|c}
\toprule
\textbf{Paradigm} & \textbf{Method} & \textbf{Stage} & \textbf{NC}$\uparrow$ & \textbf{DAC}$\uparrow$ & \textbf{DDC}$\uparrow$ & \textbf{TLC}$\uparrow$ & \textbf{EP}$\uparrow$ & \textbf{TTC}$\uparrow$ & \textbf{LK}$\uparrow$ & \textbf{HC}$\uparrow$ & \textbf{EC}$\uparrow$ & \textbf{Score}$\uparrow$ & \textbf{EPDMS}$\uparrow$ \\

\midrule
\multicolumn{14}{c}{\textit{Regression-based Planner}} \\ 
\midrule
\multirow{6}{*}{LTF \cite{chitta2022transfuser}} & \multirow{2}{*}{Base} & S1 & 96.2 & 79.6 & \textbf{99.1} & \textbf{99.6} & \textbf{84.1} & 95.1 & 94.2 & 97.6 & 79.1 & 68.9 & \multirow{2}{*}{25.2} \\
 & & S2 & 77.9 & 70.2 & 84.3 & 98.1 & 85.1 & 75.8 & 45.4 & 95.8 & \textbf{76.0} & 37.6 & \\ \cmidrule{2-14}
 & \multirow{2}{*}{DR} & S1 & 96.6 & \textbf{82.4} & 98.8 & 99.3 & 83.6 & \textbf{96.4} & \textbf{95.8} & 97.6 & 71.6 & \textbf{70.9} & \multirow{2}{*}{30.2} \\
 & & S2 & \textbf{81.9} & \textbf{73.6} & \textbf{85.9} & \textbf{98.6} & 84.7 & \textbf{78.0} & 47.2 & 96.3 & 69.3 & \textbf{41.8} & \\ \cmidrule{2-14}
 & \multirow{2}{*}{DINO (Ours)} & S1 & \textbf{97.7} & 81.3 & 98.8 & \textbf{99.6} & 83.5 & 95.8 & 93.8 & \textbf{97.8} & \textbf{80.0} & 70.6 & \multirow{2}{*}{\textbf{30.4}} \\
 & & S2 & 80.5 & 72.5 & 84.2 & 98.1 & \textbf{85.4} & 76.6 & \textbf{48.4} & \textbf{97.3} & 75.1 & 40.5 & \\

\midrule
\multicolumn{14}{c}{\textit{Diffusion-based Planner}} \\
\midrule
\multirow{6}{*}{DiffusionDrive \cite{liao2025diffusiondrive}} & \multirow{2}{*}{Base} & S1 & 96.4 & 80.0 & \textbf{99.1} & \textbf{99.3} & 83.7 & 95.1 & 93.1 & \textbf{97.8} & 77.3 & 68.2 & \multirow{2}{*}{27.1} \\
 & & S2 & 79.2 & 70.1 & 84.0 & 98.3 & 85.4 & 76.6 & 46.2 & \textbf{96.7} & \textbf{74.7} & 39.9 & \\ \cmidrule{2-14}
 & \multirow{2}{*}{DR} & S1 & 96.9 & \textbf{85.1} & 98.8 & \textbf{99.3} & \textbf{84.2} & 96.0 & \textbf{96.0} & \textbf{97.8} & 78.7 & \textbf{74.3} & \multirow{2}{*}{29.5} \\
 & & S2 & 79.4 & 70.1 & 83.7 & \textbf{98.5} & \textbf{86.6} & 76.8 & 46.6 & 96.1 & 72.7 & 39.2 & \\ \cmidrule{2-14}
 & \multirow{2}{*}{DINO (Ours)} & S1 & \textbf{97.4} & 82.7 & 99.0 & \textbf{99.3} & 83.5 & \textbf{96.4} & 94.7 & 97.6 & \textbf{80.4} & 71.8 & \multirow{2}{*}{\textbf{33.1}} \\
 & & S2 & \textbf{81.3} & \textbf{76.1} & \textbf{87.9} & 98.2 & 86.2 & \textbf{79.2} & \textbf{48.7} & 96.6 & 72.0 & \textbf{45.3} & \\

\midrule
\multicolumn{14}{c}{\textit{Scoring-based Planner}} \\
\midrule
\multirow{6}{*}{GTRS-Dense \cite{li2025generalized}} & \multirow{2}{*}{Base} & S1 & \textbf{98.9} & 95.1 & 98.9 & 99.1 & 73.9 & \textbf{98.9} & 94.7 & 96.9 & 41.8 & 76.5 & \multirow{2}{*}{45.4} \\
 & & S2 & \textbf{92.5} & 88.2 & \textbf{94.3} & 98.6 & 70.9 & \textbf{90.9} & \textbf{53.9} & 94.9 & 52.8 & 58.7 & \\ \cmidrule{2-14}
 & \multirow{2}{*}{DR} & S1 & 98.4 & 96.2 & 99.0 & \textbf{100.0} & 75.7 & 98.7 & 95.1 & 96.7 & \textbf{48.9} & \textbf{79.7} & \multirow{2}{*}{\textbf{46.8}} \\
 & & S2 & 89.2 & 89.7 & 93.5 & \textbf{99.0} & \textbf{74.8} & 87.5 & 52.0 & \textbf{97.3} & \textbf{56.8} & 58.4 & \\ \cmidrule{2-14}
 & \multirow{2}{*}{DINO (Ours)} & S1 & 97.3 & \textbf{96.7} & \textbf{99.6} & 99.8 & \textbf{76.1} & 97.8 & \textbf{96.2} & \textbf{97.6} & 42.7 & 78.5 & \multirow{2}{*}{46.7} \\
 & & S2 & 90.1 & \textbf{90.6} & 94.1 & 98.6 & \textbf{74.8} & 87.9 & 53.1 & 97.2 & 53.0 & \textbf{59.9} & \\
 
\bottomrule
\end{tabular}
}
\end{table*}

\subsection{Extended Qualitative Comparison across Planning Paradigms}
We provide extended qualitative visualizations for the diffusion-based (DiffusionDrive \cite{liao2025diffusiondrive}) and scoring-based (GTRS \cite{li2025generalized}) paradigms to further validate the versatility of the Constant Eye interface. Although the main manuscript demonstrates results for the regression-based LTF \cite{chitta2022transfuser}, these additional comparisons highlight how our DINO-based interface addresses paradigm-specific failure modes under appearance OOD. \par
\begin{figure*}[t!]
    \centering
    \includegraphics[width=\textwidth]{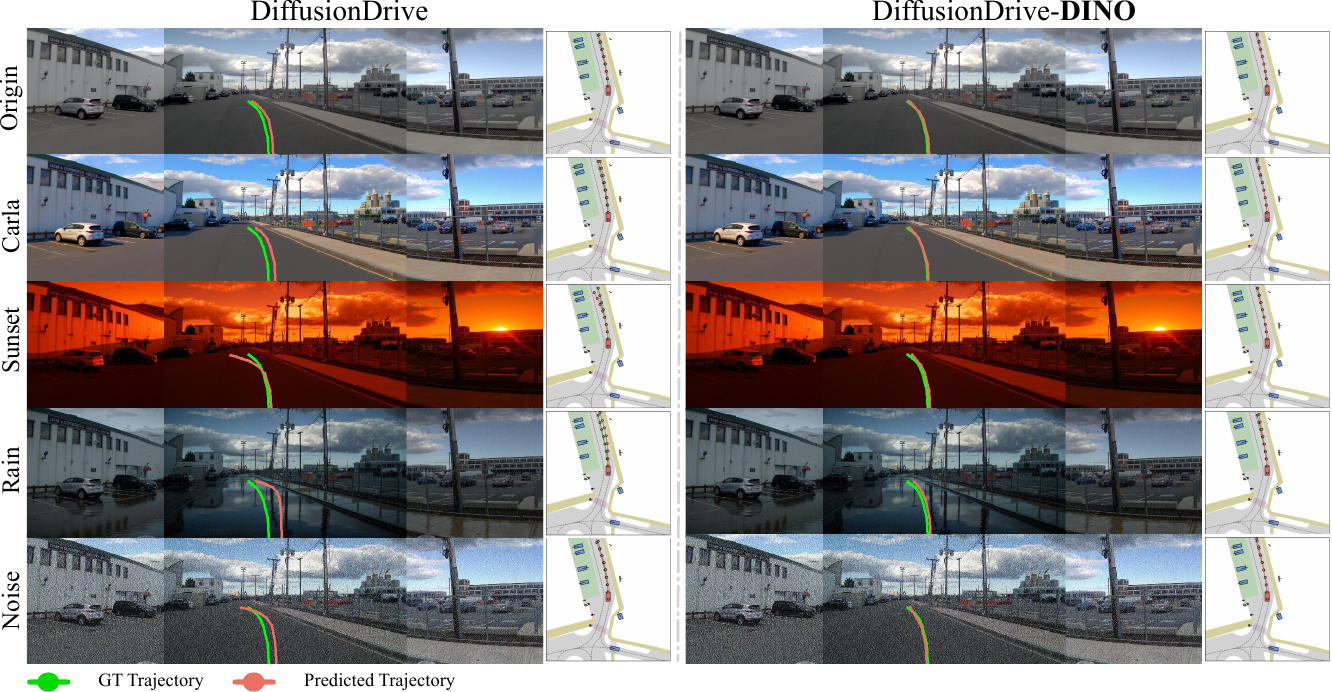}
    \caption{\textbf{Qualitative comparison between DiffusionDrive and DiffusionDrive-DINO.} We visualize the planning performance of the DiffusionDrive baseline (left) and DiffusionDrive-DINO (right) across diverse domains.}
    \label{fig:supp_diffusiondrive_comparison}
\end{figure*}

\begin{figure*}[t!]
    \centering
    \includegraphics[width=\textwidth]{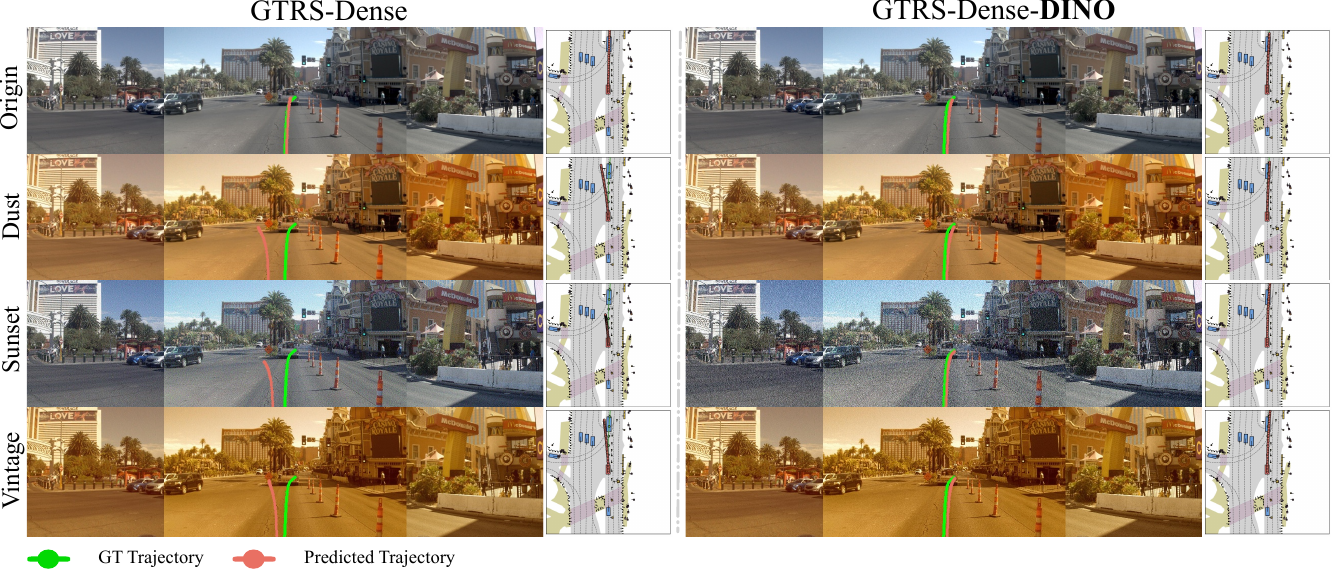}
    \caption{\textbf{Qualitative comparison between GTRS and GTRS-DINO.} We visualize the planning performance of the GTRS baseline (left) and GTRS-DINO (right) across diverse domains.}
    \label{fig:supp_gtrs_comparison}
\end{figure*}

 
As shown in \cref{fig:supp_diffusiondrive_comparison}, vanilla DiffusionDrive exhibits severe sensitivity to visual domain shifts. When subjected to varying OOD appearances, the diffusion process generates highly inconsistent trajectory distributions that significantly deviate from the expert demonstrations. This suggests that the latent representations of the standard backbone are easily corrupted by appearance shifts. In contrast, DiffusionDrive-DINO consistently decodes stable trajectory distributions that closely align with the ground truth across all tested appearance domains, demonstrating that the DINO-based interface provides a style-invariant semantic anchor that enables the diffusion process to remain robust. Similarly, as shown in \cref{fig:supp_gtrs_comparison}, vanilla GTRS fails to maintain a reliable evaluation criterion when encountering OOD appearance shifts. Due to the corruption of visual features, the baseline model often assigns misleadingly high confidence scores to unsafe candidate trajectories. In contrast, GTRS-DINO leverages the robust semantic embeddings from the foundation model to maintain a clear understanding of the drivable space. Even under severe style shifts, GTRS-DINO accurately identifies and selects the best trajectory.

\subsection{Limitations and Future Work}
\noindent
\textbf{Focus on Vision-Centric Robustness and Future Multi-modal Extensions.}
Our study primarily focuses on vision-centric planning, a paradigm that is rapidly gaining prominence due to its scalability and cost-effectiveness in real-world deployment. By utilizing a multi-camera configuration, \benchmarkname provides a foundational and highly valuable benchmark for assessing how appearance-based OOD shifts impact planning logic. However, we recognize that practical autonomous driving systems can further benefit from multi-modal integration. Future work could explore the synergy between our appearance-robust interface and other modalities, such as LiDAR or Radar, to investigate multi-modal OOD robustness.

\smallskip 
\noindent
\textbf{Open-loop Evaluation and Evolving Towards Close-loop Interaction.} The current evaluation is conducted in an open-loop manner, which serves as an essential and rigorous protocol for isolating the planner's decision-making quality from the compounding noise of control systems and vehicle dynamics. This setup allows for a clear, large-scale assessment of how visual domain shifts directly affect trajectory distribution. Building upon these valuable insights, a natural and promising next step is to extend our framework into a closed-loop setting. This would involve integrating the Constant Eye interface into interactive simulators to further validate its long-horizon stability and its ability to handle bi-directional interactions with dynamic traffic participants in real-time.

\end{document}